%% file: main_arxiv.tex
\documentclass[11pt, a4paper, logo, copyright]{hail_lab}

\usepackage[utf8]{inputenc} % allow utf-8 input
\usepackage[T1]{fontenc}    % use 8-bit T1 fonts
\usepackage{url}            % simple URL typesetting
\usepackage{booktabs}       % professional-quality tables
\usepackage{amsfonts}       % blackboard math symbols
\usepackage{nicefrac}       % compact symbols for 1/2, etc.
\usepackage{microtype}      % microtypography
\usepackage{xcolor}         % colors

\usepackage{microtype}
\usepackage{graphicx}
\usepackage{subfig}
\usepackage{booktabs}
\usepackage{amsmath,bm}
\usepackage{amsthm}
\usepackage{natbib}
\usepackage{enumerate}
\usepackage{wrapfig}
\usepackage{color, soul}
\usepackage{psfrag}
\usepackage{adjustbox}
\usepackage{xspace}
\usepackage{amssymb}
\usepackage{multirow}
\usepackage{afterpage}
\usepackage{colortbl}
\usepackage{float}
\usepackage{textcomp}
\usepackage{siunitx}
\usepackage{mdframed}
\usepackage{makecell}
\usepackage[super]{nth}

\usepackage{enumitem}
\usepackage{bbding}
\usepackage{pifont}
\usepackage{bbm}
\usepackage{xcolor}
\usepackage{pdflscape}
\usepackage{theoremref}
\usepackage{pifont}
\usepackage{adjustbox}
\usepackage{xcolor}
\usepackage{mathtools}
\usepackage{subcaption}
\usepackage{pbox}
\usepackage{longtable}
\usepackage{listings}

\colorlet{darkgreen}{green!65!black}
\colorlet{darkblue}{blue!75!black}
\colorlet{darkred}{red!80!black}
\definecolor{statistical}{HTML}{8c564b}
\definecolor{structural}{HTML}{0070C0}
\definecolor{semantic}{HTML}{008080}
\definecolor{yellow}{HTML}{f7c600}
\definecolor{lightblue}{HTML}{0071bc}
\definecolor{lightgreen}{HTML}{39b54a}
\definecolor{deemph}{gray}{0.55}

\definecolor{baselinecolor}{gray}{.95}
\definecolor{graycolor}{gray}{.95}
\newcommand{\grayrow}{\rowcolor[gray]{.95}}

\usepackage[pagebackref=false, breaklinks=true, colorlinks, citecolor=lightblue, urlcolor=lightblue, bookmarks=false]{hyperref}

\newlength\savewidth
\newcolumntype{x}[1]{>{\centering\arraybackslash}p{#1pt}}
\newcolumntype{y}[1]{>{\raggedright\arraybackslash}p{#1pt}}
\newcolumntype{z}[1]{>{\raggedleft\arraybackslash}p{#1pt}}

\newcommand{\ours}{\texttt{SensorGen}\xspace}

\newcommand{\semantictosignal}{\mbox{Semantic-to-Signal}\xspace}
\newcommand{\interpextrap}{\mbox{Interpolation \& Extrapolation}\xspace}
\newcommand{\translation}{\mbox{Translation}\xspace}
\newcommand{\signalediting}{\mbox{Editing}\xspace}

\newcommand{\numtasksetting}{\mbox{14}\xspace}

\newcommand{\numdataset}{\mbox{7}\xspace}
\newcommand{\nummodality}{\mbox{12}\xspace}
\newcommand{\numtaskcategory}{\mbox{4}\xspace}
\newcommand{\numdomain}{\mbox{4}\xspace}
\newcommand{\datasetmimiciecg}{MIMIC-IV ECG\xspace}
\newcommand{\datasetppgdalia}{PPG-DaLiA\xspace}
\newcommand{\datasetshhs}{SHHS\xspace}
\newcommand{\datasetphymer}{PhyMER\xspace}
\newcommand{\datasetcapturetwentyfour}{CAPTURE-24\xspace}
\newcommand{\datasetvitaldb}{VitalDB\xspace}
\newcommand{\datasetmetabonet}{Metabonet\xspace}

\newcommand{\domained}{Emergency Department\xspace}
\newcommand{\domaindailylife}{Daily Life\xspace}
\newcommand{\domainoperationroom}{Operation Room\xspace}
\newcommand{\domainlab}{Lab Study\xspace}

\newcommand{\dit}{{DiT}\xspace}
\newcommand{\sit}{{SiT}\xspace}
\newcommand{\fractalgen}{{FractalGen}\xspace}
\newcommand{\tarflow}{{TarFlow}\xspace}
\newcommand{\mar}{{MAR}\xspace}
\newcommand{\imagen}{{Imagen}\xspace}

\definecolor{textgreen}{RGB}{57, 172, 57}
\definecolor{textred}{RGB}{200, 10, 10}
\definecolor{boxyellow}{HTML}{FAF5E6}
\definecolor{frameyellow}{HTML}{B7950B}
\definecolor{boxpurple}{HTML}{F4EFF6}
\definecolor{framepurple}{HTML}{6C3483}
\definecolor{boxblue}{HTML}{EEF4F8}
\definecolor{frameblue}{HTML}{2874A6}
\definecolor{boxgray}{HTML}{F0F2F3}
\definecolor{framegray}{HTML}{5D6D7E}
\definecolor{boxgreen}{HTML}{EAFaf1}
\definecolor{framegreen}{HTML}{196F3D}
\usepackage{pifont} 
\newcommand{\cmark}{\textcolor{textgreen}{\ding{51}}}
\newcommand{\xmark}{\textcolor{textred}{\ding{55}}}

\usepackage[most]{tcolorbox}
\tcbset{
    center title,
    left=0pt,
    right=0pt,
    top=0pt,
    bottom=0pt,
    colframe=black,
    colback=gray!10,
    width=\linewidth,
    enlarge left by=0mm,
    boxsep=5pt,
    boxrule=1pt,
    arc=0pt,outer arc=0pt,
}

\newtcolorbox{promptbox}[1][]{
    enhanced,
    colback=white,
    colframe=black,
    fonttitle=\bfseries,
    title=Prompt,
    attach boxed title to top left={xshift=10pt, yshift*=-\tcboxedtitleheight/2},
    boxed title style={colback=black},
    top=12pt, bottom=10pt, left=10pt, right=10pt,
    #1
}

\tcbset{
    fancybox/.style={
        enhanced,
        breakable,
        arc=1mm,
        outer arc=1mm,
        boxrule=0pt,
        leftrule=1mm,
        toptitle=0.5mm,
        bottomtitle=0.5mm,
        fonttitle=\bfseries\sffamily\small,
        fontupper=\scriptsize,
        coltitle=white,
        drop shadow
    }
}

\newtcolorbox{thoughtbox}{
    fancybox,
    colback=boxyellow,
    colframe=frameyellow,
    coltitle=black,
    title=Thought
}
\newtcolorbox{userbox}{
    fancybox,
    colback=boxpurple,
    colframe=framepurple,
    title=User
}
\newtcolorbox{agentbox}{
    fancybox,
    colback=boxblue,
    colframe=frameblue,
    title=Agent
}
\newtcolorbox{outputbox}{
    fancybox,
    colback=boxgray, 
    colframe=framegray,
    coltitle=black,
    title=Execution Output
}
\newtcolorbox{solutionbox}{
    fancybox,
    colback=boxgreen,
    colframe=framegreen,
    title=Solution
}

\definecolor{codegreen}{rgb}{0.0, 0.5, 0.0}
\definecolor{codegray}{rgb}{0.4, 0.4, 0.4}
\definecolor{codepurple}{rgb}{0.50, 0, 0.50}
\definecolor{backcolour}{rgb}{0.97, 0.97, 0.97}

\lstdefinestyle{mystyle}{
    backgroundcolor=\color{backcolour},
    commentstyle=\color{codegreen},
    keywordstyle=\color{magenta},
    stringstyle=\color{codepurple},
    basicstyle=\ttfamily\scriptsize, % Readable size
    breakatwhitespace=false,
    breaklines=true,
    captionpos=b,
    keepspaces=true,
    numbers=none,              % Changed: explicit instruction to hide numbers
    showspaces=false,
    showstringspaces=false,
    showtabs=false,
    tabsize=2,
    frame=single,
    rulecolor=\color{black!10}, % Very subtle gray frame
    frameround=fttt,            % Rounded corners on the code frame
    upquote=true
}
\newcommand{\finding}[2]{%
  \begin{tcolorbox}[
    enhanced,
    breakable,
    width=\linewidth,
    colback=white!90!gray,
    colframe=white,        % no visible frame
    boxrule=0pt,           % remove frame line
    arc=5pt,
    boxsep=5pt,
    left=5pt,right=5pt,top=3pt,bottom=3pt,
    before skip=0.6\baselineskip,
    after skip=0.6\baselineskip,
  ]
  \noindent\textbf{\textit{Finding~#1:}}~#2
  \end{tcolorbox}%
}
\newcommand{\takeaway}[2]{%
  \begin{tcolorbox}[
    enhanced,
    breakable,
    width=\linewidth,
    colback=white!90!gray,
    colframe=white,        % no visible frame
    boxrule=0pt,           % remove frame line
    arc=5pt,
    boxsep=5pt,
    left=5pt,right=5pt,top=3pt,bottom=3pt,
    before skip=0.6\baselineskip,
    after skip=0.6\baselineskip,
  ]
  \noindent\textbf{\textit{Takeaway~#1:}}~#2
  \end{tcolorbox}%
}

\title{
Signal or Noise? Understanding Generative Models for Real-World Sensor Time Series
}

\author[1$*$]{Zitao Shuai}
\author[1$*$]{Zongzhe Xu}
\author[2$*$]{Yuntian Wu}
\author[1]{Sirui Li}
\author[3]{Tianhong Li}
\author[1$\dagger$]{Yuzhe Yang}

\affil[1]{University of California, Los Angeles}
\affil[2]{Carnegie Mellon University}
\affil[3]{Massachusetts Institute of Technology}
\correspondingauthor{$^*$Equal contribution. $^\dagger$Correspondence to: yuzhey@ucla.edu.}

\datasetlink{https://huggingface.co/yang-ai-lab/SensorGen}
\codelink{https://github.com/yang-ai-lab/SensorGen}
\projecturl{https://yang-ai-lab.github.io/sensor-gen}

\input{sections/0_abstract}

\begin{document}

\maketitle
\newenvironment{Itemize}{
    \begin{itemize}[leftmargin=*]
    \setlength{\itemsep}{0pt}
    \setlength{\topsep}{0pt}
    \setlength{\partopsep}{0pt}
    \setlength{\parskip}{1pt}}
{\end{itemize}}
\setlength{\leftmargini}{9pt}

%%%%%%%%%%%%%%%%%%%%%%%%%%%%%%%%%%%%%%%%%%%%%%%%%%%%%%%%%%%%%%%%%%%%%%%%%%%%%%%
% MAIN BODY
%%%%%%%%%%%%%%%%%%%%%%%%%%%%%%%%%%%%%%%%%%%%%%%%%%%%%%%%%%%%%%%%%%%%%%%%%%%%%%%
%\input{sections/0_abstract}
\input{sections/1_intro}
\input{sections/2_related_work}
\input{sections/3_methods}
\input{sections/4_results}
\input{sections/5_discussion}
\input{sections/6_conclusion}

%%%%%%%%%%%%%%%%%%%%%%%%%%%%%%%%%%%%%%%%%%%%%%%%%%%%%%%%%%%%%%%%%%%%%%%%%%%%%%%
% ACKNOWLEDGMENTS (hidden in anonymous submission mode)
%%%%%%%%%%%%%%%%%%%%%%%%%%%%%%%%%%%%%%%%%%%%%%%%%%%%%%%%%%%%%%%%%%%%%%%%%%%%%%%
\section*{Acknowledgments}
We gratefully acknowledge the support by the Amazon Science Hub, the NVIDIA Academic Grant Program, and UCLA DataX. Any opinions, findings, conclusions, or recommendations expressed in this material are those of the author(s) and do not necessarily reflect the views of the funders.

% \clearpage
\bibliography{ref}
\bibliographystyle{plain}

%%%%%%%%%%%%%%%%%%%%%%%%%%%%%%%%%%%%%%%%%%%%%%%%%%%%%%%%%%%%
\newpage
\appendix
\input{sections/7_appendix}

\end{document}

%% file: sections/0_abstract.tex
\begin{abstract}
\vspace{-7pt}

Generative models have changed how machine learning represents complex data distributions, especially in language and vision, yet many real-world systems are observed instead as continuous, high-dimensional, and noisy \textit{sensor time series}.
Existing generative modeling of sensor data, however, remains fragmented across modalities, datasets, and task formulations, limiting a systematic understanding of \textit{when}, \textit{how}, and \textit{why} generative models succeed or fail in real-world settings. To address this gap, we introduce \ours, a large-scale study of sensor-signal generation spanning \numtasksetting settings across \numdomain domains, \numdataset datasets, and \nummodality signal modalities. Leveraging \ours, we systematically evaluate generative models from five major families and uncover three key findings: 
(1) flow-matching models provide strong overall performance across most settings; 
(2) signal properties matter, with demographic covariates improving longitudinal generation and time-frequency modeling improving high-frequency signal generation; and
(3) generated signals have practical utility beyond visual realism, with scaling improving generation quality and synthetic data improving downstream performance.
Together, \ours establishes a broader understanding of design choices, evaluation protocols, and failure modes in real-world sensor data generation.

\end{abstract}

%% file: sections/1_intro.tex
\vspace{-5pt}
\section{Introduction}
\label{sec:intro}
\vspace{-3pt}

Generative models have reshaped how machine learning imagines data: learning not only to recognize patterns, but to sample, complete, and represent complex distributions \cite{peebles2023scalable,ho2020denoising}.
Yet, much of the real world is not expressed as language or images, but as continuous, noisy, high-dimensional \textit{\textbf{sensor time series}} that record physiology \cite{xu2026sleeplm}, behavior \cite{chan2024capture}, environments \cite{nguyen2023climax}, and machines \cite{khazatsky2024droid}. 
This makes sensor time series a central data regime for modern machine learning, but extending generative modeling advances to sensor signals is non-trivial.
Unlike language or natural images, sensor signals are collected across diverse settings and modalities, and vary widely in sampling frequency, time span, sequence length, channel structure, and physical semantics.

Existing studies for sensor time series \cite{lai2025diffusets,ding2024self,ibtehaz2022ppg2abp}, however, remain highly \textit{fragmented} across modalities, datasets, and task formulations. Current sensor generation methods are often designed for specific applications, with modeling paradigms, architectures, and evaluation protocols tailored to a particular signal type or task.
While effective in their target settings, these dedicated designs provide limited evidence about whether the modeling choices generalize across heterogeneous sensor generative problems.
As a result, the field lacks a systematic understanding of the factors that shape generation quality across sensor signals.
This motivates a central question:
\vspace{-3pt}
\begin{center}
\emph{When, how, and why do generative models succeed or fail\\ across diverse sensor time series generation settings?}
\end{center}
\vspace{-3pt}

\begin{figure}[!t]
\centering
\includegraphics[width=0.99\linewidth]{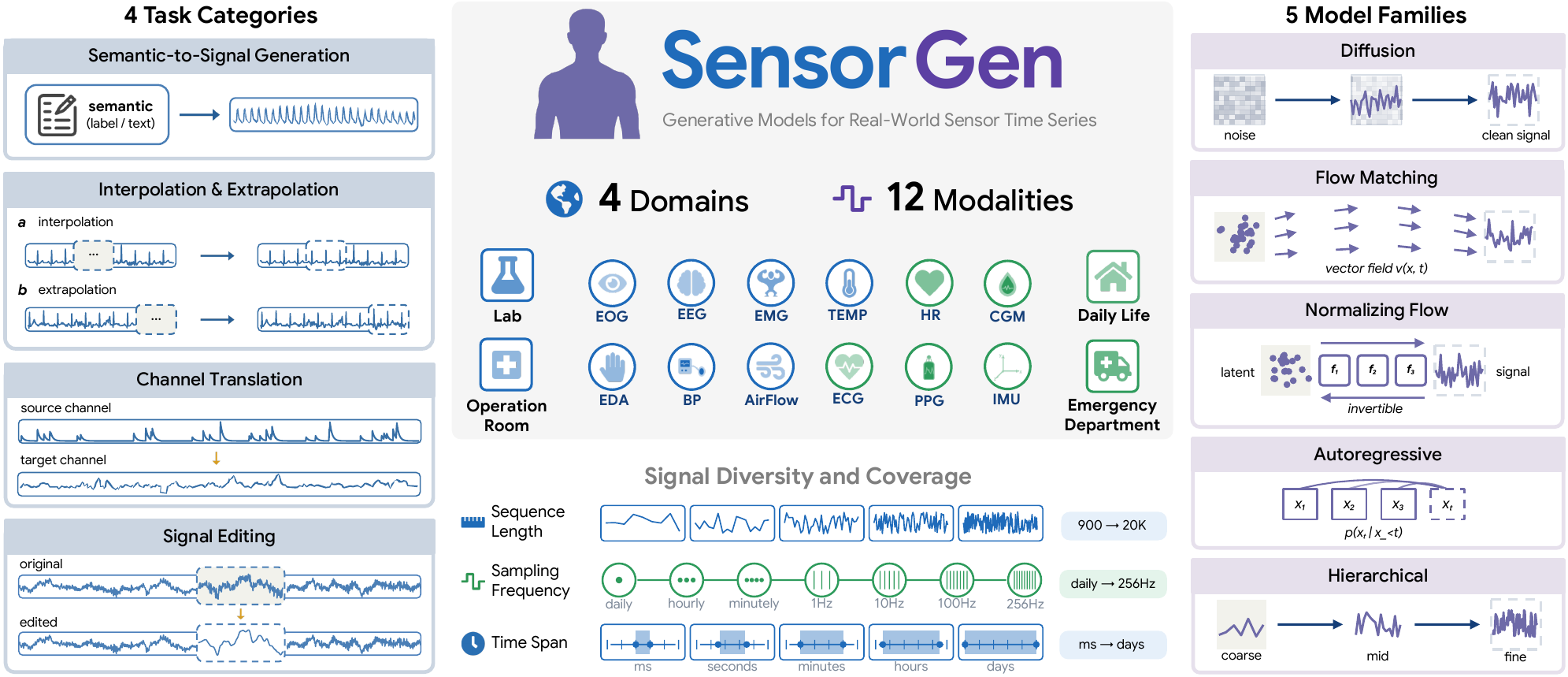}
\vspace{-2pt}
\caption{\small{
\textbf{Overview of \ours.} We present a large-scale study of real-world sensor time-series generation, spanning \numtasksetting settings across \numdomain domains, \numdataset datasets, and \nummodality signal modalities. \ours establishes to date the broadest coverage of sequence length, frequency, and time span. More details are in Appendix \ref{appendix:task_formulation}.
}}
\label{fig:teaser_figure}
\vspace{-10pt}
\end{figure}

To answer this question, we introduce \ours, a fully open and large-scale exploration of generative modeling for real-world sensor time series. Rather than introducing another task-specific generator, \ours is designed to disentangle how model families, signal properties, task formulations, scale, and evaluation protocols jointly shape generation quality (Fig.~\ref{fig:teaser_figure}):
\ding{182} \textbf{Task coverage.} We construct practically meaningful generation tasks from real-world sensor datasets and standardize them into a unified pipeline for data processing, task construction, model training, and evaluation.
\ding{183} \textbf{Sensor diversity.} \ours spans \numtasksetting generation settings across \numdomain domains, \numdataset datasets, and \nummodality signal modalities, covering diverse sampling frequencies, sequence lengths, and time spans.
\ding{184} \textbf{Generation settings.} We organize these settings into \numtaskcategory major categories: \textit{semantic-to-signal generation}, \textit{sensor interpolation and extrapolation}, \textit{channel translation}, and \textit{signal editing}.
\ding{185} \textbf{Model families.} We evaluate six representative models from five major generative families, including diffusion models, flow-matching models, autoregressive models, normalizing-flow models, and hierarchical models.

Using \ours, we reveal intriguing lessons for building generative models for sensor signals.
First, \textit{model family matters}: flow-matching models consistently provide strong performance across most settings, suggesting a strong baseline for heterogeneous sensor generation.
Second, \textit{sensor-property-aware design matters}: demographic covariates improve longitudinal generation, while time-frequency modeling improves high-resolution signal generation.
Third, \textit{generated signals are useful beyond realism}: scaling generally improves generation quality, and synthetic data improves downstream task performance under data scarcity. 
Beyond these findings, we further analyze design choices, evaluation approaches, and failure modes, providing a broader understanding of key gaps and
directions for future research. Our contributions are as follows:
\vspace{-5pt}
\begin{Itemize}
\item We introduce \ours, the first open large-scale study of sensor time-series generation, spanning \numtasksetting settings across \numdomain domains, \numdataset datasets, and \nummodality signal modalities, with broad coverage of sequence length, sampling frequency, and time span.
\vspace{2pt}
\item We establish a unified exploration framework that consolidates fragmented sensor-generation tasks, modeling paradigms, and evaluation protocols, covering \numtaskcategory generation categories and five representative generative model families.
\vspace{2pt}
\item We verify that flow-matching serves as strong baselines across diverse settings, while signal-aware designs, including demographic conditioning and time-frequency modeling, improve generation for longitudinal and high-frequency data.
\vspace{2pt}
\item We demonstrate that generated signals are useful beyond visual realism: scaling improves generation quality, and synthetic data improves downstream performance under data scarcity.
\vspace{2pt}
\item We provide extensive analyses of design choices, evaluation metrics, and failure modes, offering practical lessons for developing and evaluating generative models for real-world sensor time series.
\end{Itemize}

%% file: sections/2_related_work.tex
\vspace{-3pt}
\section{Related Work}
\vspace{-5pt}

\textbf{Modern Generative Modeling.} Generative models have become a central approach for learning and sampling complex data distributions. Recent advances span several representative families, including diffusion models \cite{peebles2023scalable,ho2020denoising}, flow-matching models \cite{ma2024sit,esser2024scaling}, autoregressive models \cite{li2024autoregressive,sun2024autoregressive}, normalizing flows \cite{zhai2024normalizing}, and hierarchical generative models \cite{saharia2022photorealistic, li2025fractal}. These families have achieved strong performance in high-fidelity image synthesis \cite{dhariwal2021diffusion} and language generation \cite{brown2020language}, making them natural candidates for studying generation beyond language and vision.
However, existing evidence is largely concentrated in these domains, leaving open which model families are most effective for sensor data generation.
Unlike language or natural images, sensor signals are continuous, noisy, temporally structured, and highly heterogeneous in sampling frequency, time span, sequence length, channel structure, and physical semantics \cite{li2026hearts}, making direct adaptation to sensor time series non-trivial.

\textbf{Generative Models for Sensor Time Series.} Recent work has leveraged generative models for practical sensor-signal tasks, including synthesizing ECG signals to improve sample diversity or augment rare conditions \cite{lai2025diffusets, hu2024personalized, bondar2025flowecg,neifar2024diffecg}, translating PPG signals into ECG or blood-pressure waveforms \cite{ding2024self, shome2024region,fang2025ppgflowecg}, generating inertial measurement unit signals \cite{oppel2025diffusion}, forecasting vital signs and wearable signals \cite{ibtehaz2022ppg2abp,chen2026versatile}, synthesizing realistic neurophysiological time series \cite{vetter2024generating}, imputing missing measurements \cite{kazijevs2023deep}, and editing or denoising physiological signals \cite{jang2026cofe}. However, existing works are often developed for specific modalities, datasets, or applications, with task-specific model designs, conditioning strategies, preprocessing pipelines, and evaluation protocols (see Table \ref{tab:related_work_comparison}).
In contrast, \ours presents an in-depth study of when, how, and why generative models succeed or fail across sensor-signal generation settings, unifying diverse tasks and representative model families under a shared protocol to analyze how model choices, signal properties, and task formulations shape generation quality.

\input{tables/related_work_comparison}

%% file: tables/related_work_comparison.tex
% Colors
\definecolor{barbg}{gray}{0.88}
\definecolor{seqcolor}{RGB}{103,158,213}
\definecolor{freqcolor}{RGB}{238,164,127}
\definecolor{timecolor}{RGB}{137,185,145}

% Optional check/cross marks
% \usepackage{pifont}
% \newcommand{\cmark}{\ding{51}}
% \newcommand{\xmark}{\ding{55}}

\newcommand{\seqbar}[3]{%
    \makebox[12mm][r]{\raisebox{0.2ex}{\tiny #3}}%
    \makebox[21mm][l]{%
        \rlap{\textcolor{barbg}{\rule{21mm}{1.5ex}}}%
        \hspace{#1mm}%
        \textcolor{seqcolor}{\rule{#2mm}{1.5ex}}%
    }%
}

\newcommand{\specbar}[3]{%
    \makebox[18mm][r]{\raisebox{0.2ex}{\tiny #3}}%
    \makebox[21mm][l]{%
        \rlap{\textcolor{barbg}{\rule{21mm}{1.5ex}}}%
        \hspace{#1mm}%
        \textcolor{freqcolor}{\rule{#2mm}{1.5ex}}%
    }%
}

\newcommand{\timebar}[3]{%
    \makebox[16mm][r]{\raisebox{0.2ex}{\tiny #3}}%
    \makebox[21mm][l]{%
        \rlap{\textcolor{barbg}{\rule{21mm}{1.5ex}}}%
        \hspace{#1mm}%
        \textcolor{timecolor}{\rule{#2mm}{1.5ex}}%
    }%
}

\newcommand{\dashbar}{%
    \makebox[18mm][r]{\raisebox{0.2ex}{\tiny --}}%
    \makebox[21mm][l]{%
        \rlap{\textcolor{barbg}{\rule{21mm}{1.5ex}}}%
    }%
}

\begin{table*}[!t]
  \centering
  \caption{\small{
  \textbf{Comparison of studies on generative modeling of sensor time series.}
    % We summarize recent studies built around modern generative models for sensor signals.
    % While prior work is typically specialized to individual settings, \ours provides a large-scale study spanning four task categories and five modern generative model families.
    }}
  \label{tab:related_work_comparison}
  \vspace{-4pt}
  \begin{adjustbox}{width=\textwidth}
  \footnotesize
  \setlength{\tabcolsep}{5pt}
  \begin{tabular}{l ccc ccccc cccc}
    \toprule[1.5pt]
    \multirow{2.5}{*}{\textbf{Study}} &
    \multirow{2.5}{*}{\textbf{\# Domain}} &
    \multirow{2.5}{*}{\textbf{\# Dataset}} &
    \multirow{2.5}{*}{\textbf{\# Modality}} &
    \multicolumn{5}{c}{\textbf{Model Family}} &
    \multicolumn{4}{c}{\textbf{Task Type}} \\
    \cmidrule(lr){5-9}\cmidrule(lr){10-13}
    & & & &
    \texttt{Diff} & \texttt{FM} & \texttt{NF} & \texttt{AR} & \texttt{Hier} &
    \texttt{S2S}  & \texttt{IntExt} & \texttt{Trans} & \texttt{Edit} \\
    \midrule
    \midrule

    % Existing works
    % Lai \textit{et al.}~\cite{lai2025diffusets} % ecg
    %   & 1 & 2 & 1
    %   & \cmark & \xmark & \xmark & \xmark & \xmark
    %   & \cmark & \xmark & \xmark & \xmark \\

    Vetter \textit{et al.}~\cite{vetter2024generating} % eeg
      & 1 & 4 & 3
      & \cmark & \xmark & \xmark & \xmark & \xmark
      & \cmark & \cmark & \xmark & \xmark \\
    Oppel \textit{et al.}~\cite{oppel2025diffusion} %imu
      & 1 & 1 & 1
      & \cmark & \xmark & \xmark & \xmark & \xmark
      & \cmark & \xmark & \xmark & \xmark \\
    Bondar \textit{et al.}~\cite{bondar2025flowecg} % ecg; flow ecg
    & 1 & 1 & 1
      & \xmark & \cmark & \xmark & \xmark & \xmark
      & \cmark & \xmark & \xmark & \xmark \\
    % Shome \textit{et al.}~\cite{shome2024region} % ppg to ecg; rddm
    % & 1 & 7 & 3
    %   & \cmark & \xmark & \xmark & \xmark & \xmark
    %   & \xmark & \xmark & \cmark & \xmark
    % \\
    Fang \textit{et al.}~\cite{fang2025ppgflowecg} % ppg to ecg; flow ppg2ecg
    & 1 & 4 & 2
      & \xmark & \cmark & \xmark & \xmark & \xmark
      & \xmark & \xmark & \cmark & \xmark
    \\
    Neifar \textit{et al.}~\cite{neifar2024diffecg} %diffecg
    & 1 & 1 & 1
      & \cmark & \xmark & \xmark & \xmark & \xmark
      & \cmark & \cmark & \xmark & \xmark
    \\
    Chen \textit{et al.}~\cite{chen2026versatile} %unicardio
    & 1 & 4 & 3
      & \cmark & \xmark & \xmark & \xmark & \xmark
      & \xmark & \cmark & \cmark & \cmark
    \\
    \midrule

    \rowcolor[gray]{.95}
    \textbf{\ours (ours)}
      & \textbf{\numdomain}
      & \textbf{\numdataset}
      & \textbf{\nummodality}
      & \textbf{\cmark}
      & \textbf{\cmark}
      & \textbf{\cmark}
      & \textbf{\cmark}
      & \textbf{\cmark}
      & \textbf{\cmark}
      & \textbf{\cmark}
      & \textbf{\cmark}
      & \textbf{\cmark} \\

    \bottomrule[1.5pt]
  \end{tabular}
  \end{adjustbox}
\vspace{4pt}
\newline
\scriptsize
  \texttt{Diff}: Diffusion.~~~\texttt{FM}: Flow-Matching.~~~\texttt{NF}: Normalizing Flow.~~~\texttt{AR}: Autoregressive.~~~\texttt{Hier}: Hierarchical Modeling.

\vspace{2pt}
  \texttt{S2S}: Semantic-to-Signal.~~~\texttt{IntExt}: Interpolation \& Extrapolation.~~~\texttt{Trans}: Translation.~~~\texttt{Edit}: Signal Editing / Inverse Problems.
\vspace{-10pt}
\end{table*}

%% file: sections/3_methods.tex
\vspace{-5pt}
\section{SensorGen}
\label{sec:benchmark}
\vspace{-4pt}

\ours bridges the fragmented landscape of sensor time-series generation into a unified foundation. Instead of building a single best generative model, we study sensor generation as a joint problem over \textit{task formulations}, \textit{sensing domains}, \textit{signal properties}, and \textit{model families}. This is essential for sensor data, where generation quality can vary substantially across modalities with different sampling frequencies, temporal spans, sequence lengths, channel structures, and physical semantics \cite{li2026hearts}. 
% Meanwhile, generative modeling paradigms developed in broader domains bring distinct generation mechanisms and assumptions about data structure, leaving their effectiveness on sensor signals underexplored.

% \ours is designed as a fully open and extensible study platform for examining these interactions. It covers \numdataset{} datasets, \numtasksetting{} task settings, and \nummodality{} signal modalities, and evaluates five major generative model families under shared task construction, training, and evaluation pipelines. Together, these components allow us to examine not only which models perform best, but also how signal properties, generative paradigms, and design choices jointly shape reliable sensor time-series generation across real-world physiological and healthcare scenarios.

% \begin{figure}
%     \centering
%     \includegraphics[width=0.6\linewidth]{figures/task_coverage_all_cropped.pdf}
%     \caption{\textbf{Task coverage of \ours.}}
%     \label{fig:task coverage}
% \end{figure}

\subsection{Data Regimes}
\label{subsec:data_overview}
\vspace{-3pt}

\begin{wrapfigure}{r}{0.45\textwidth}
\centering
\vspace{-18pt}
\includegraphics[width=0.35\textwidth]{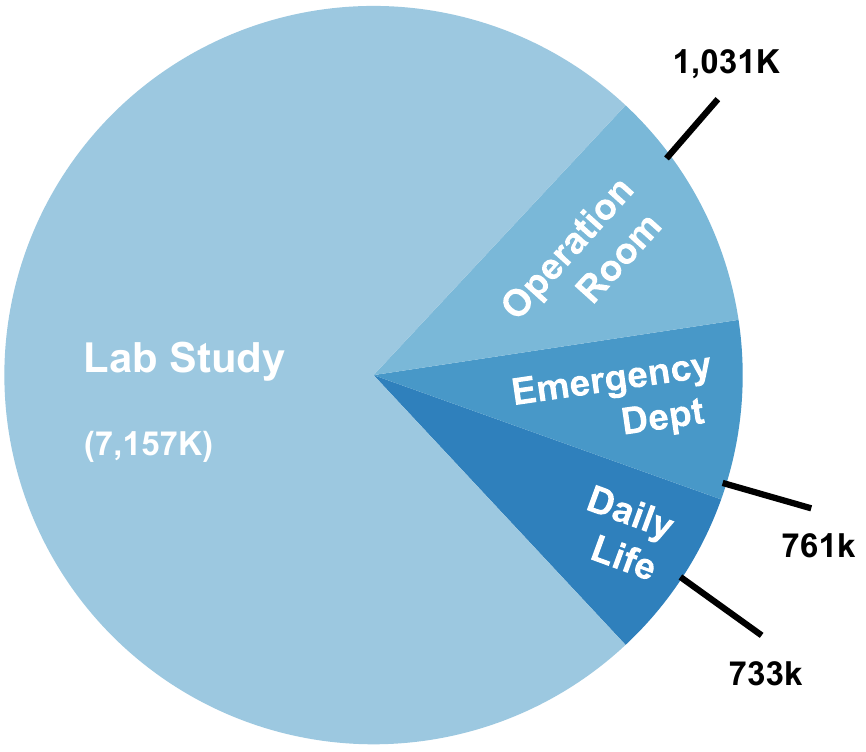}
\caption{\textbf{Domain coverage of \ours.}}
\label{fig:domain coverage}
\vspace{-6pt}
\end{wrapfigure}

\textbf{Dataset diversity and domain coverage.} 
A central design of \ours is to organize sensor generation around the major real-world regimes in which sensor data are collected and used, rather than simply aggregating datasets by availability.
We therefore define four broad, distinct, and representative collection scenarios: \textit{emergency-department} monitoring, \textit{daily-life} sensing, \textit{laboratory} studies, and \textit{operation-room} monitoring.
We then instantiate these scenarios with \numdataset{} widely used public datasets, spanning applications in everyday human behavioral sensing, health monitoring, and clinical care, and capturing signals related to physiological, emotional, behavioral, and metabolic states.
This design provides a broad view of real-world sensing contexts:
\vspace{-4pt}
\begin{Itemize}
    \item \textit{\textbf{\domained}}: Emergency-department ECG datasets, including MIMIC-IV ECG \cite{gow2023mimic}, provide clinical 12-lead electrocardiogram recordings paired with diagnostic reports.
\vspace{3pt}
    \item \textit{\textbf{\domaindailylife}}: Daily-life sensing datasets span CAPTURE-24 \cite{chan2024capture}, PPG-DaLiA \cite{reiss2019deep}, and Metabonet \cite{wolff2026metabonet}, provide free-living wrist accelerometry, wearable physiological measurements such as PPG and ECG, and longitudinal metabolic traces such as CGM, insulin, and carbohydrate intake.
\vspace{3pt}
    \item \textit{\textbf{\domainlab}}: Signals collected from laboratory monitoring environments and sleep-study datasets, including PhyMER \cite{pant2023phymer} and SHHS \cite{zhang2018national,quan1997sleep}, provide controlled affective-state recordings and overnight polysomnography, covering EEG, ECG, EMG, EOG, and peripheral physiological signals.
\vspace{3pt}
    \item \textit{\textbf{\domainoperationroom}}: VitalDB \cite{lee2022vitaldb} provides intra-operative physiological waveforms, medication-rate records, arterial blood pressure, PPG, ECG, and NIBP-derived measurements.
\end{Itemize}

\textbf{Broad coverage of sensor modalities.}
\ours includes \nummodality{} signal modalities spanning bioelectrical recordings, optical and hemodynamic measurements, wearable motion signals, intra-operative waveforms, intervention records, and longitudinal metabolic traces.
These modalities differ not only in physical origin, channel structure, and physiological semantics, but also in how they are sampled, measured, and observed in practice.
As a result, \ours spans diverse signal properties, including \textit{sequence length}, \textit{sampling frequency}, and \textit{time span}.
Across tasks, sequence lengths range from hundreds to tens of thousands of steps, sampling frequencies span from one sample every five minutes to 256 Hz, and observation windows range from seconds to seven days.

% \vspace{-3pt}
\subsection{Task Taxonomy}
\label{sec:task_overview}
% \vspace{-3pt}

\textbf{Design Principles and Formulation.}
We design the task taxonomy of \ours around real sensing bottlenecks where generative modeling is practically useful.
To instantiate each setting, we use three criteria that address whether the task is valid, valuable, and generatively modelable: 
\ding{182} \textit{Evidence-Grounded Validity}, ensuring support from clinical practice, knowledge, or prior empirical findings; 
\ding{183} \textit{Application Value}, targeting a practical data bottleneck or deployment need; and 
\ding{184} \textit{Generative Suitability}, requiring temporal resolution and span to contain meaningful signal structure.

\begin{wrapfigure}{r}{0.45\textwidth}
\vspace{-1pt}
    \centering
    \includegraphics[width=0.45\textwidth]{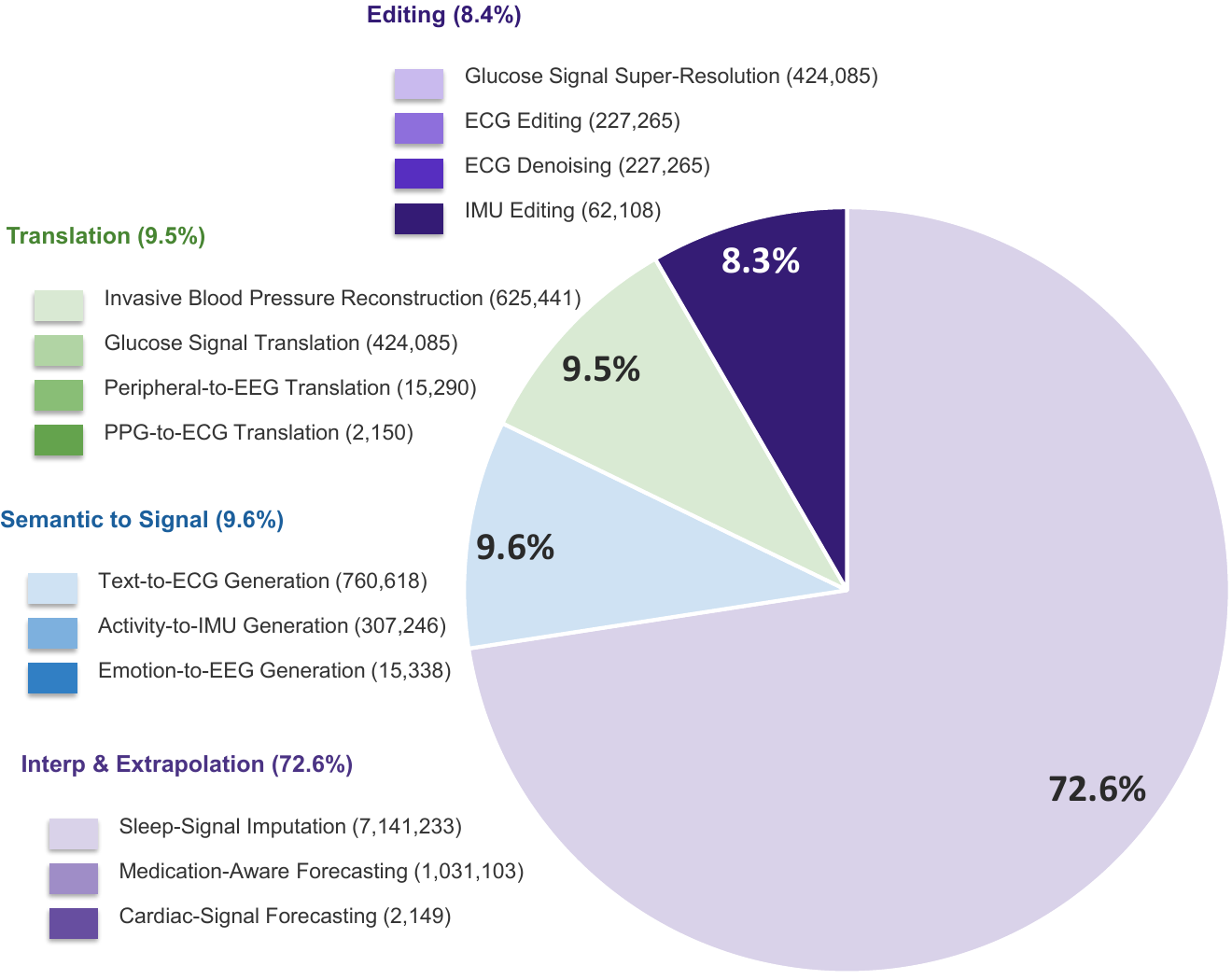}
    \caption{\textbf{Task coverage of \ours.}}
\vspace{-8pt}
    \label{fig:task coverage}
\end{wrapfigure}

\begin{figure}
    \centering
    \includegraphics[width=1\linewidth]{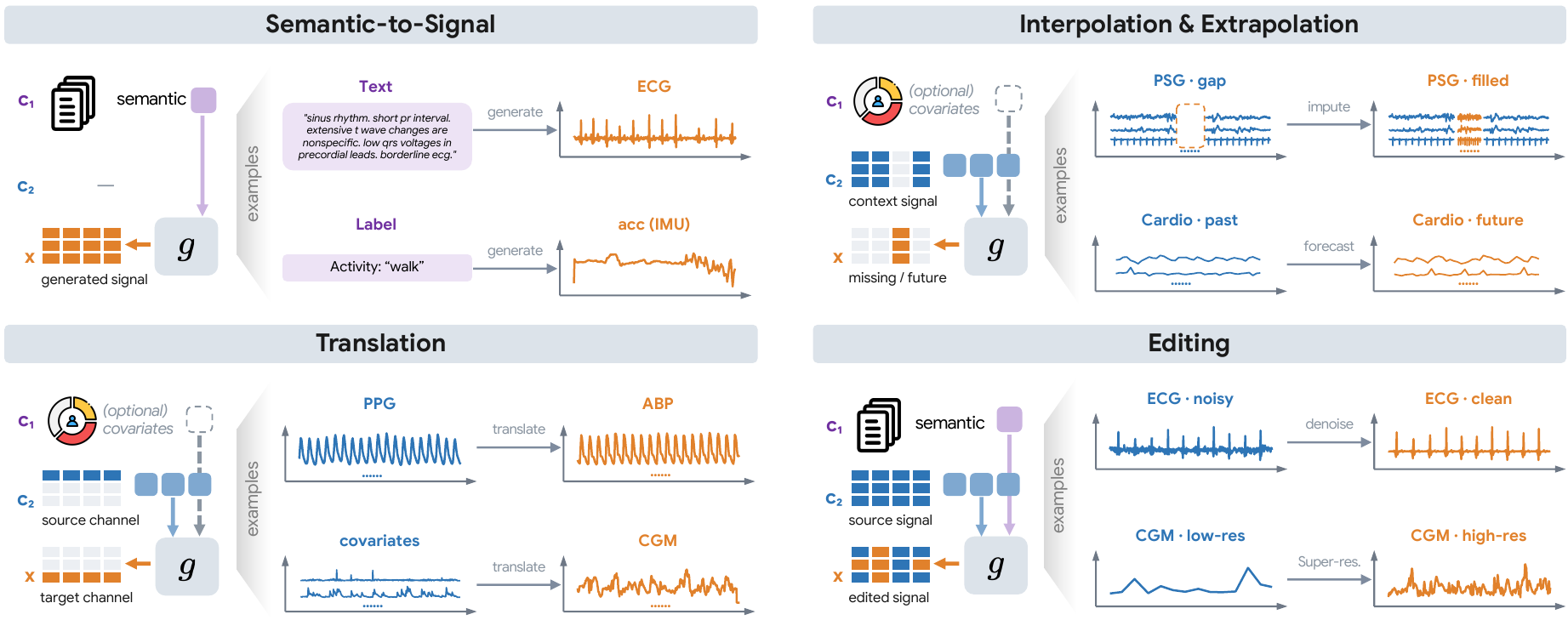}
    \caption{\textbf{Illustration of task categories in \ours.} For each category, we show the task-level data flow (left) and representative generation settings with example channels (right).}
    \vspace{-8pt}
    \label{fig:task category}
\end{figure}

Following these principles, we construct \numtaskcategory{} generation categories:
\ding{182} \textit{\semantictosignal}, addressing the scarcity of diverse and semantically annotated signal samples;
\ding{183} \textit{Interpolation \& Extrapolation}, targeting future prediction and reconstruction of temporally incomplete signals;
\ding{184} \textit{\translation}, reconstructing missing or hard-to-acquire channels from accessible sources; and
\ding{185} \textit{\signalediting}, enabling constrained restoration or counterfactual modification of existing recordings.
Fig. \ref{fig:task coverage} summarizes its coverage breadth.

We formulate all tasks as conditional generation problems. 
Let $\mathbf{x}\in\mathbb{R}^{C\times T}$ denote signal observations.
Here, $c_1$ includes text, labels, or metadata, while $c_2$ includes historical signals, source channels, or other observed signal context. 
Each task can be written as $\hat{\mathbf{x}}\sim p_\theta\left(\mathbf{x}\mid\mathcal{C}\right)$, where $\mathcal{C}\subseteq\{c_1,c_2\}$ is the set of available conditions. 
Fig.~\ref{fig:task category} shows the modeling settings of each task category, and we also summarize the condition-target protocol for each category below, with detailed formulations and dataset-specific settings in Appendix \ref{appendix:task_formulation}.

\vspace{-7pt}
\begin{Itemize}

\item \textit{\textbf{\semantictosignal.}} 
Given a semantic condition $c_1$, models generate a target signal $\mathbf{x}$ that is realistic and semantically consistent. Task families include:
\begin{itemize}[nosep]
    \item \textit{Text-to-Signal Generation:} Synthesizing physiological signals from text descriptions.
    \item \textit{Label-Conditioned Generation:} Generating sensor signals guided by discrete labels.
\end{itemize}

\vspace{3pt}
\item \textit{\textbf{\interpextrap.}}
Given temporal context $c_2$, and optionally auxiliary condition $c_1$, models generate unobserved signal segments $\mathbf{x}$. Task families include:
\begin{itemize}[nosep]
    \item \textit{Future Forecasting:} Predicting future segments from historical context.
    \item \textit{Intervention-Conditioned Forecasting:} Predicting future physiological responses conditioned on history and intervention variables.
    \item \textit{Temporal Imputation:} Reconstructing missing segments from surrounding observations.
\end{itemize}

\vspace{3pt}
\item \textit{\textbf{\translation.}}
Given source channels or proxy measurements $c_2$, and optionally auxiliary condition $c_1$, models generate target channels $\mathbf{x}$ from the same physiological episode. Task families include:
\begin{itemize}[nosep]
    \item \textit{Signal Translation:} Generating a target modality from a source modality using correlations.
    \item \textit{Proxy-to-Target Reconstruction:} Synthesizing clinically valuable hard-to-access signals from accessible proxies.
\end{itemize}

\vspace{3pt}
\item \textit{\textbf{\signalediting.}}
Given an observed signal, and optional guidance $c_1$, models generate a restored, refined, or edited target signal $\mathbf{x}$ while preserving most of the non-targeted attributes. We consider both training-free and training-based variants of this setting, including:
\begin{itemize}[nosep]
    \item \textit{Super-Resolution:} Solving an inverse problem of recovering high-resolution trajectories from low-resolution observations.
    \item \textit{Denoising:} Reconstructing clean signals from recordings disturbed with random noise.
    \item \textit{Semantic Signal Editing:} Modifying a signal toward a target semantic condition while preserving non-targeted semantics.
\end{itemize}
\end{Itemize}
\vspace{-10pt}

\vspace{-3pt}
\subsection{Model Families}
\label{subsec:model_families}
\vspace{-3pt}

Our goal is to compare mainstream modern generative model families under a unified sensor-generation setting, rather than to build a task-specific model for each dataset or application. We select model families based on three criteria: \ding{182} \textit{Contemporary relevance}, prioritizing paradigms that have become strong baselines for high-fidelity generation \cite{dhariwal2021diffusion, peebles2023scalable}; \ding{183} \textit{Mechanistic diversity}, covering distinct generation mechanisms and modeling assumptions; and \ding{184} \textit{Task alignment}, favoring approaches that naturally match the challenges in our task taxonomy \cite{song2020score, saharia2022photorealistic}. The selected model families are summarized below, with implementation details provided in Appendix~\ref{appendix:model_detail}. For each family, we pick the most representative models as an instantiation.

\vspace{-3pt}
\begin{Itemize}
    \item \textbf{Diffusion Models.} 
    Diffusion models generate samples through iterative denoising from noise to data. We instantiate this family with \textsc{DiT} \cite{ho2020denoising,peebles2023scalable}.
\vspace{3pt}
    \item \textbf{Flow Matching.} 
    Flow-matching models learn a continuous transport process from noise to data. We instantiate this family with \textsc{SiT} \cite{ma2024sit}.
\vspace{3pt}
    \item \textbf{Autoregressive Models.} 
    Autoregressive models generate tokens sequentially by conditioning each step on previously generated tokens. We instantiate this family with \textsc{MAR} \cite{li2024autoregressive}.
\vspace{3pt}
    \item \textbf{Normalizing Flows.} 
    Normalizing flows model data through invertible likelihood-based transformations. We instantiate this family with \textsc{TarFlow} \cite{zhai2024normalizing}.
\vspace{3pt}
    \item \textbf{Hierarchical Models.} 
    Hierarchical models generate samples through multi-scale or coarse-to-fine synthesis. We instantiate this family with \textsc{FractalGen} \cite{li2025fractal} and \textsc{Imagen} \cite{saharia2022photorealistic}.
\end{Itemize}

%% file: sections/4_results.tex
\input{tables/main_translation_interpextrap}

\vspace{-7pt}
\section{Main Results}
\vspace{-5pt}

\textbf{Data setup.}
We use a unified preprocessing pipeline that standardizes heterogeneous sensor datasets into a shared data representation, enabling consistent task construction, model training, and evaluation across domains and modalities. 
Unless otherwise specified, each loaded window is normalized independently with min--max scaling to \(\left[-1, 1\right]\), which stabilizes training for generative modeling. 
Detailed preprocessing and data-loading procedures are provided in Appendix~\ref{app:datasets}.

\textbf{Training Setup.}
We benchmark representative generative model families for sensor time series. All methods share unified task definitions, preprocessing pipelines, and data splits, and are trained from scratch for each task setting with matched batch sizes, training budgets, and comparable model capacities when applicable. We train each run for up to 50K steps or until convergence, and do not over-optimize hyperparameters and training techniques for a specific method. Additional implementation details are provided in Appendix \ref{appendix:training_settings}.

\textbf{Evaluation Protocol.}
For each method and setting, we randomly generate 1,024 samples from held-out test conditions and report task-specific metrics (see Appendix \ref{appendix:task details}). For semantic-to-signal generation, we compute distribution-level metrics in the feature space of MIRA \cite{li2025mira}, a representation model pre-trained on diverse healthcare time-series data. For paired generation tasks, we report sample-level metrics, including MSE, MAE, PSNR, spectrogram-domain MSE (SMSE), and SSIM \cite{wang2004image}. We follow~\cite{shome2024region} to calculate these metrics based on the normalized signal.

\input{tables/main_semantic_to_signal_superres}

\vspace{-5pt}
\subsection{Which generative paradigm transfers best to sensor time series?}
\vspace{-3pt}

\textbf{Model family matters, with flow matching emerging as the strongest default across tasks.}
We first compare representative generative models across all task categories. For each metric, we aggregate results within each task category and report category-level averages. Overall, \sit provides the strongest performance, although no single method dominates every setting. As shown in Table \ref{tab:translation_aggregate} and \ref{tab:interpextrap_aggregate}, \sit achieves the best aggregate performance on both \translation{} and \interpextrap{} tasks, consistently outperforming other methods across the reported metrics.

\input{tables/main_editing}
For \semantictosignal{}, Table \ref{tab:semantic_to_signal_aggregate} shows that \sit also achieves better overall distribution-level fidelity. In contrast, \signalediting{} tasks are more setting-dependent: Table \ref{tab:signal_editing_aggregate} shows that \imagen remains competitive on training-required editing, while Table \ref{tab:editing_sample_level_aggregate} confirms \sit performs best in training-free editing settings. 
Full setting-level results are provided in Appendix \ref{sec:appendix:detailed results}.

These results suggest that flow matching is the most reliable default choice for sensor-signal generation in our study. At the same time, the remaining methods show task-dependent strengths, indicating that modern generative paradigms do not transfer uniformly to sensor signals. Their inductive biases may align differently with sensor-specific requirements, including temporal dynamics, spectral structure, and conditional consistency. We next analyze how these signal properties shape model behavior.

\takeaway{1}{Flow-matching models provide a strong baseline for sensor time series generation.}

\begin{figure}
\centering
\includegraphics[width=1\linewidth]{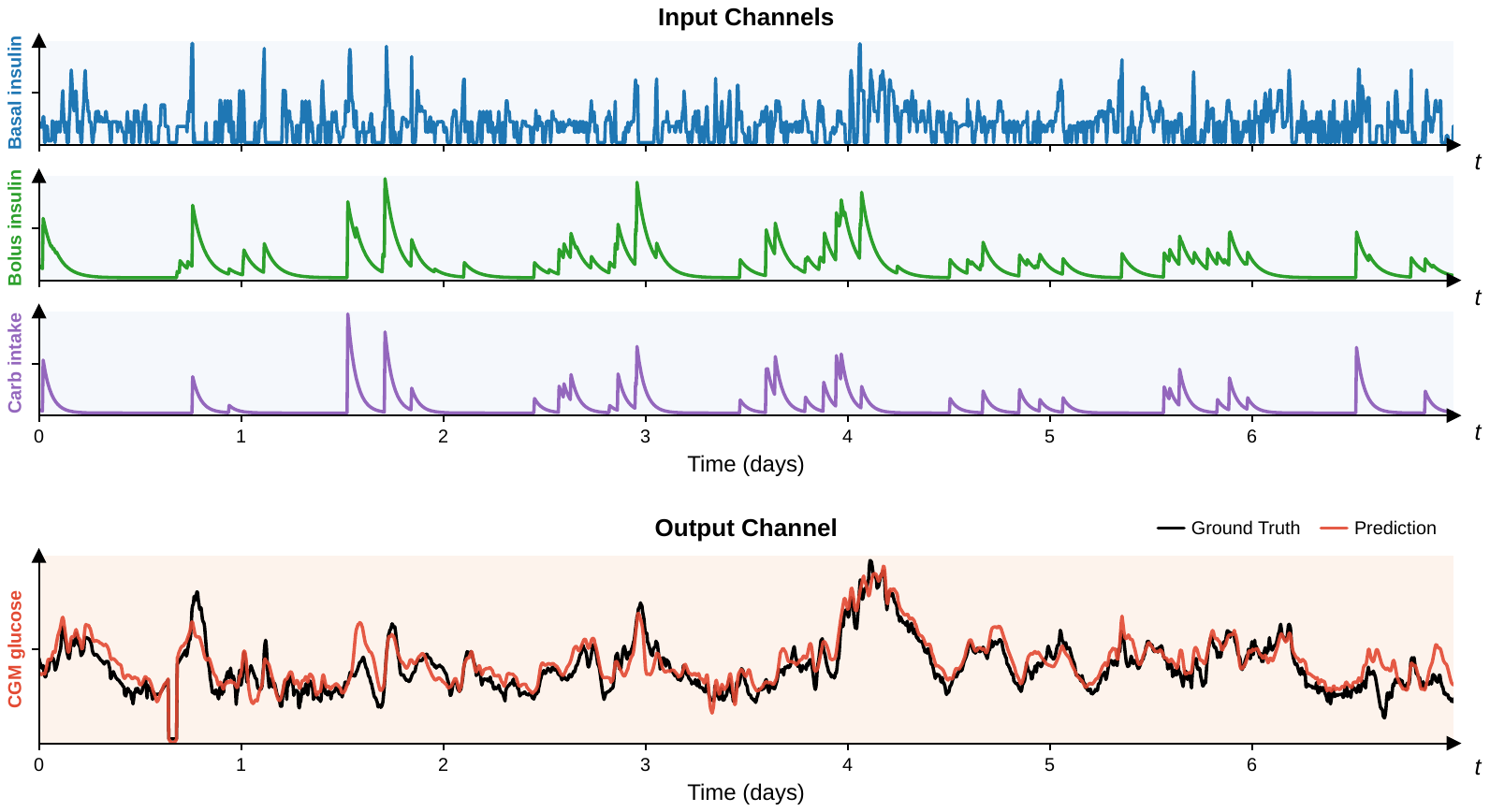}
\vspace{-17pt}
\caption{\small{\textbf{Visualization of generated signals.}
Example from the glucose signal translation task. \textbf{Top:} Input signals. \textbf{Bottom:} Ground truth and the generated target signal.
More examples are in Appendix \ref{appendix:qualitative result}.
}}
\label{fig:main viz}
\vspace{-10pt}
\end{figure}

\textbf{Qualitative Results.} In Fig.~\ref{fig:main viz}, we show a representative generation example on glucose signal translation. More visualizations are provided in Appendix \ref{appendix:qualitative result}. Across settings, generated samples capture recognizable signal structure rather than degenerating into noise, supporting the quantitative finding that modern generative models can produce realistic sensor time series under diverse conditions.

% for main results' table here, I think semantic to signal, we can do 3 setting per row. others may need adjustment

% semantic to signal: 3 metric * 3 + 1 admin; 1 large table
% translation: 2 large tables
% 5 metric * 2 + 1 admin
% forecasting: 1 large table + 1 small table
% 5 metric * 2 + 1 admin
% 5 + 1

% editing:
% 2 small tables
% (5 metrics + 2 admin) x 3 for editing & denoising
% (5 metric + 1 admin) x 1 for CGM superresolution
% \input{tables/main_translation}
% \input{tables/main_forecast}
% \input{tables/main_imputation_superres}
% \input{tables/main_translation_2}
% \input{tables/main_semantic2signal}
% \input{tables/main_editing}
% \input{tables/main_forecast}

% aggregation logics:
% pretraining settings: 4 category --> 4 tables
% inference settings: 3 category --> 1 small table

\subsection{What context matters for generating long-horizon sensor data?}

\input{tables/ablation_demo}
\textbf{Subject context helps longitudinal generation only when encoded carefully.}
We next examine which factors matter for generating longitudinal sensor signals over extended time spans. Motivated by strong subject-level heterogeneity in long-term physiological patterns \cite{yang2022artificial, zhang2025sensorlm}, we study whether demographic covariates provide useful global context for generation.

As shown in Table~\ref{tab:glucose_translation_ablation}, directly using raw demographic attributes substantially degrades performance compared with using no demographic information. In contrast, dataset-level normalization makes demographic conditioning beneficial, outperforming the no-demographic baseline. These results suggest that subject-level metadata can improve longitudinal generation, but only when encoded in a properly normalized form.

\takeaway{2}{Normalized demographic covariates improve longitudinal sensor-signal generation.}

\subsection{Do generative models scale to longer sensor sequences?}

\begin{figure}
\centering
\includegraphics[width=1\linewidth]{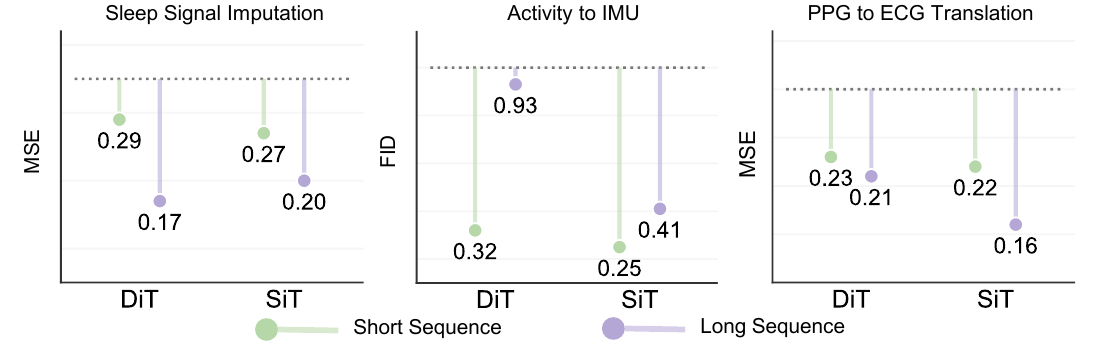}
\caption{\textbf{Comparison of long-sequence generation.} Models struggle with long signals in semantic-to-signal generation, but outperform on imputation and translation when sensor context is available.}
\label{fig:long sequence}
\vspace{-15pt}
\end{figure}

\textbf{Generative models can handle long sensor sequences when target-signal context is available.}
We next examine how sequence length affects generation quality, since long sensor sequences require models to maintain temporal consistency over extended horizons. To this end, we evaluate paired short- and long-sequence settings: \textit{SHHS} sleep-signal imputation and \textit{Capture-24} activity-conditioned IMU generation use sequence lengths of 1,920 \textit{vs.} 19,200 and 900 \textit{vs.} 9,000, while \textit{PPG-to-ECG translation} uses sequence lengths of 1,000 \textit{vs.} 6,000. For comparable evaluation, different from settings in the main table, we split each generated longer signal into slices matching the short-sequence duration, compute metrics on each slice, and report the average.

As shown in Fig. \ref{fig:long sequence}, generation quality degrades as sequence length increases for semantic-to-signal generation, but remains stable in the channel translation and imputation setting. This contrast suggests that long-sequence generation is more tractable when target-signal context is available: surrounding observed segments can anchor the missing region and reduce the burden of generating long-range structure from condition information alone.

\takeaway{3}{Target-signal context reduces degradation in long-sequence generation.}

\subsection{How should models capture high-frequency signal semantics?}

\textbf{High-frequency signal semantics are difficult to capture, while time-frequency modeling improves fidelity.}
We next examine high-frequency signal generation, where semantic fidelity depends not only on waveform morphology but also on spectral structure. As shown in Table \ref{tab:semantic_to_signal_distribution_level}, for high-frequency signals such as EEG, all models achieve relatively low recall despite high precision and realistic morphology in qualitative examples, as shown in Appendix \ref{appendix:qualitative result}. 
This suggests that current generative models can produce plausible high-frequency waveforms, but still struggle to cover the semantic and spectral variability of EEG signals.

To test whether explicit time-frequency structure can mitigate this limitation, we conduct an ablation on the emotion-to-EEG setting. We extend the input with the spectrogram to verify whether additional frequency-domain guidance leads to performance improvement. Specifically, we transform raw EEG waveforms into spectrograms using the short-time Fourier transform (STFT), and use the spectrogram as a time-frequency conditioning signal. As shown in Fig. \ref{fig:spectrogram_help}, incorporating this representation improves the fidelity of high-frequency channel generation.

\takeaway{4}{High-frequency signal generation benefits from time-frequency modeling.}

%% file: tables/main_translation_interpextrap.tex
\begin{table*}[!t]
    \centering

    \begin{minipage}[t]{0.49\textwidth}
    \centering
    \caption{\small{{Results on \textit{\translation{}}.}}}
    \label{tab:translation_aggregate}
    \vspace{-2pt}
    \begin{adjustbox}{width=\linewidth}
    \footnotesize
    \setlength{\tabcolsep}{3pt}
    \begin{tabular}{l *{5}{r}}
      \toprule[1.5pt]
      \multirow{2.5}{*}{\textbf{Model}} &
      \multicolumn{5}{c}{\textbf{\translation{}}} \\
      \cmidrule(lr){2-6}
      & MSE$^\downarrow$ & MAE$^\downarrow$ & SMSE$^\downarrow$ & PSNR$^\uparrow$ & SSIM$^\uparrow$ \\
      \midrule
      \midrule

      \dit
        & 0.224 & 0.343 & 0.044 & 13.33 & 0.258 \\
\grayrow
      \sit
        & \textbf{0.214} & \textbf{0.336} & \textbf{0.041} & \textbf{13.61} & \textbf{0.259} \\
      \fractalgen
        & 0.357 & 0.454 & 0.099 & 10.98 & 0.101 \\
      \tarflow
        & 0.382 & 0.482 & 0.097 & 10.54 & 0.060 \\
      \mar
        & 0.387 & 0.483 & 0.096 & 10.48 & 0.091 \\

      \bottomrule[1.5pt]
    \end{tabular}
    \end{adjustbox}
    \vspace{-3mm}
    \end{minipage}
    \hfill
    \begin{minipage}[t]{0.49\textwidth}
    \centering
    \caption{\small{{Results on \textit{Interpolation \& Extrapolation}.}}}
    \label{tab:interpextrap_aggregate}
    \vspace{-2pt}
    \begin{adjustbox}{width=\linewidth}
    \footnotesize
    \setlength{\tabcolsep}{3pt}
    \begin{tabular}{l *{5}{r}}
      \toprule[1.5pt]
      \multirow{2.5}{*}{\textbf{Model}} &
      \multicolumn{5}{c}{\textbf{\interpextrap{}}} \\
      \cmidrule(lr){2-6}
      & MSE$^\downarrow$ & MAE$^\downarrow$ & SMSE$^\downarrow$ & PSNR$^\uparrow$ & SSIM$^\uparrow$ \\
      \midrule
      \midrule

      \dit
        & 0.335 & 0.448 & 0.096 & 11.48 & 0.086 \\
\grayrow
      \sit
        & \textbf{0.252} & \textbf{0.384} & \textbf{0.064} & \textbf{12.53} & \textbf{0.129} \\
      \fractalgen
        & 0.345 & 0.413 & 0.070 & 11.47 & 0.109 \\
      \tarflow
        & 0.353 & 0.450 & 0.083 & 11.22 & 0.071 \\
      \mar
        & 0.351 & 0.431 & 0.101 & 11.35 & 0.116 \\

      \bottomrule[1.5pt]
    \end{tabular}
    \end{adjustbox}
    \vspace{-6pt}
    \end{minipage}

\end{table*}

%% file: tables/main_semantic_to_signal_superres.tex
% Version 1: 5 + 66
\begin{table*}[!t]
    \centering

    \begin{minipage}[t]{0.41\textwidth}
    \centering
    \caption{\small{Results on \textit{Semantic-to-Signal}.}}
    \label{tab:semantic_to_signal_aggregate}
    \vspace{-2pt}
    \begin{adjustbox}{width=\linewidth}
    \footnotesize
    \setlength{\tabcolsep}{5pt}
    \begin{tabular}{l *{3}{r}}
      \toprule[1.5pt]
      \multirow{2.5}{*}{\textbf{Model}} &
      \multicolumn{3}{c}{\textbf{\semantictosignal{}}} \\
      \cmidrule(lr){2-4}
      & FID$^\downarrow$ & Precision$^\uparrow$ & Recall$^\uparrow$ \\
      \midrule
      \midrule

      \dit
        & \textbf{2.05} & 0.887 & 0.363 \\
\grayrow
      \sit
        & 2.23 & \textbf{0.915} & \textbf{0.413} \\

      \fractalgen
        & 7.27 & 0.632 & 0.073 \\

      \tarflow
        & 4.26 & 0.617 & 0.111 \\

      \mar
        & 7.76 & 0.747 & 0.158 \\

      \imagen
        & 2.51 & 0.734 & 0.346 \\

      \bottomrule[1.5pt]
    \end{tabular}
    \end{adjustbox}
    \vspace{-3mm}
    \end{minipage}
    \hfill
    \begin{minipage}[t]{0.57\textwidth}
    \centering
    \caption{\small{Results on \textit{Signal \signalediting{}}.}}
    \label{tab:signal_editing_aggregate}
    \vspace{-2pt}
    \begin{adjustbox}{width=\linewidth}
    \footnotesize
    \setlength{\tabcolsep}{5pt}
    \begin{tabular}{l *{5}{r}}
      \toprule[1.5pt]
      \multirow{2.5}{*}{\textbf{Model}} &
      \multicolumn{5}{c}{\textbf{\signalediting{}}} \\
      \cmidrule(lr){2-6}
      & MSE$^\downarrow$ & MAE$^\downarrow$ & SMSE$^\downarrow$ & PSNR$^\uparrow$ & SSIM$^\uparrow$ \\
      \midrule
      \midrule

      \dit
        & 0.059 & 0.151 & 0.032 & 22.05 & 0.339 \\

      \sit
        & 0.134 & 0.260 & 0.088 & 19.35 & 0.276 \\

      \fractalgen
        & 0.127 & 0.256 & 0.042 & 15.79 & 0.314 \\

      \tarflow
        & 0.034 & 0.144 & 0.008 & 21.55 & 0.370 \\

      \mar
        & 0.266 & 0.393 & 0.121 & 12.04 & 0.118 \\
\grayrow
      \imagen
        & \textbf{0.008} & \textbf{0.064} & \textbf{0.002} & \textbf{27.83} &\textbf{ 0.699} \\

      \bottomrule[1.5pt]
    \end{tabular}
    \end{adjustbox}
    \vspace{-4pt}
    \end{minipage}

\end{table*}

%% file: tables/main_editing.tex
\begin{wraptable}{r}{0.45\linewidth}
\vspace{-6pt}
    \centering
    \caption{\small{Results on training-free \textit{Signal \signalediting{}}.}}
    \label{tab:editing_sample_level_aggregate}
    \vspace{-2pt}
    \begin{adjustbox}{width=\linewidth}
    \footnotesize
    \setlength{\tabcolsep}{4pt}
    \begin{tabular}{l *{5}{r}}
      \toprule[1.5pt]
      \multirow{2.5}{*}{\textbf{Model}} &
      \multicolumn{5}{c}{\textbf{\signalediting{} (Training-Free)}} \\
      \cmidrule(lr){2-6}
      & MSE$^\downarrow$ & MAE$^\downarrow$ & SMSE$^\downarrow$ & PSNR$^\uparrow$ & SSIM$^\uparrow$ \\
      \midrule
      \midrule

      \dit
        & 0.160 & 0.266 & 0.044 & 16.74 & 0.299 \\
\grayrow
      \sit
        & \textbf{0.155} & \textbf{0.250} & \textbf{0.043} & \textbf{18.48} & \textbf{0.342} \\

      \bottomrule[1.5pt]
    \end{tabular}
    \end{adjustbox}
\vspace{-5pt}
\end{wraptable}

%% file: tables/ablation_demo.tex
\begin{wraptable}{r}{0.52\linewidth}
  \centering
  \caption{\small{\textbf{Comparison of demographic covariate encoding strategies.} We report glucose signal translation results.}}
  \label{tab:glucose_translation_ablation}
  \vspace{-2pt}
  \begin{adjustbox}{width=\linewidth}
  \footnotesize
  \setlength{\tabcolsep}{4pt}
  \begin{tabular}{l l *{5}{r}}
    \toprule[1.5pt]
    \textbf{Model} &
    \textbf{Encoding} & MSE$^\downarrow$ & MAE$^\downarrow$ & SMSE$^\downarrow$ & PSNR$^\uparrow$ & SSIM$^\uparrow$ \\
    \midrule
    \midrule

    \dit & Raw
      & 0.767 & 0.744 & 0.578 & 7.57 & 0.017 \\
\grayrow
    \dit & Z-score
      & \textbf{0.097} & \textbf{0.223} & \textbf{0.032} & \textbf{17.38} & 0.374 \\

    \dit & No demo
      & 0.106 & 0.230 & 0.032 & 17.12 & \textbf{0.388} \\

    \midrule

    \sit & Raw
      & 0.272 & 0.420 & 0.158 & 11.90 & 0.020 \\
\grayrow
    \sit & Z-score
      & \textbf{0.091} & \textbf{0.216} & \textbf{0.030} & \textbf{17.70} & \textbf{0.400} \\

    \sit & No demo
      & 0.101 & 0.228 & 0.032 & 17.32 & 0.371 \\

    \bottomrule[1.5pt]
  \end{tabular}
    \vspace{-15pt}
  \end{adjustbox}
\end{wraptable}

%% file: sections/5_discussion.tex
\section{Analysis}
\subsection{Are generated sensor signals useful beyond visual realism?}
% (1) syn data augment supervised task
% (2) syn data augment SSL pretraining

\begin{figure}
    \centering
    \includegraphics[width=1\linewidth]{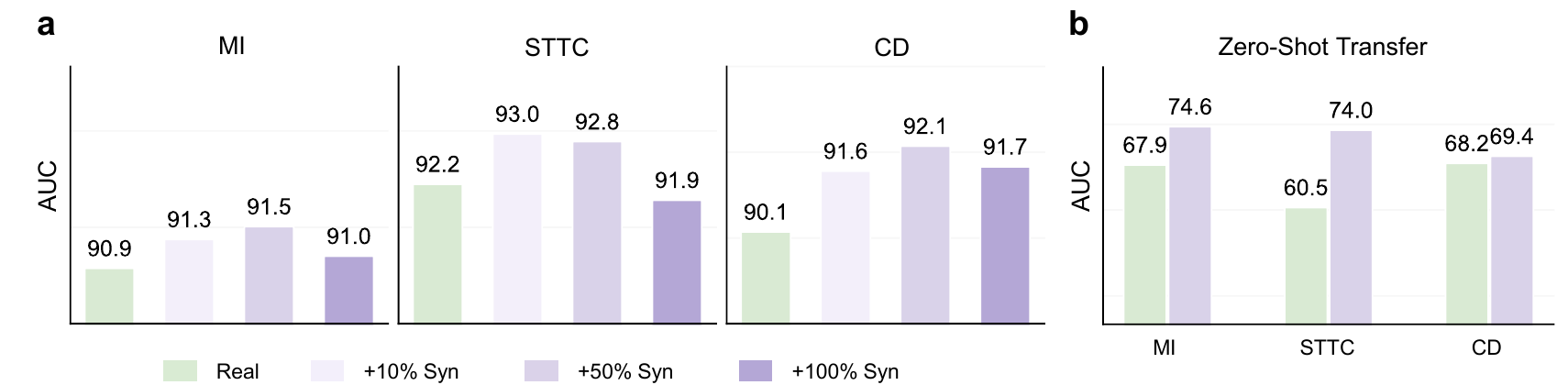}
    \caption{\small{\textbf{Downstream utility of generated signals.}
    With ECGs synthesized by a text-to-ECG generator, we evaluate their utility in two downstream settings.
    \textbf{(a)} Moderate synthetic augmentation improves disease classification performance, whereas excessive synthetic data degrades it. \textbf{(b)} Adding synthetic data improves zero-shot transfer in ECG-text representation learning.}}
    \label{fig:syn data help sup}
\vspace{-15pt}
\end{figure}

\textbf{Synthetic sensor signals benefit self-supervised pre-training and supervised learning.}
We evaluate whether generated signals provide utility beyond sample realism by using them to augment self-supervised representation learning and supervised disease prediction.
For self-supervised learning, we follow the ECG-text pre-training setting in \cite{wang2025token}, comparing encoders pretrained with real data alone versus a mixture of real and synthetic data.
We transfer the resulting models to PTB-XL \cite{wagner2020ptb} and evaluate zero-shot classification.
As shown in Fig.~\ref{fig:syn data help sup}, adding synthetic data improves the downstream performance of Myocardial Infarction (MI)
, ST/T Change (STTC)
, and Conduction Disturbance (CD) detection tasks
, indicating that generated signals can augment pre-training data and improve representation quality.

For supervised learning, we train ResNet classifiers on PTB-XL disease diagnosis tasks. We augment the real training set with labeled synthetic ECG signals at three ratios, corresponding to 10\%, 50\%, and 100\% of the real-data size.
As shown in Fig. \ref{fig:syn data help sup}, 50\% and 10\% synthetic augmentation achieve better performance; adding more synthetic data degrades performance, but still often outperforms the real-data-only baseline.
This suggests that moderate synthetic augmentation helps address data scarcity, whereas excessive synthetic data may amplify repeated patterns or generator artifacts.

\finding{1}{Synthetic sensor signals provide downstream utility through data augmentation.}

\subsection{How do design choices affect generated signal quality?}

\begin{figure*}[!t]
    \centering

\begin{minipage}[t]{0.32\textwidth}
    \centering
    \includegraphics[width=\linewidth]{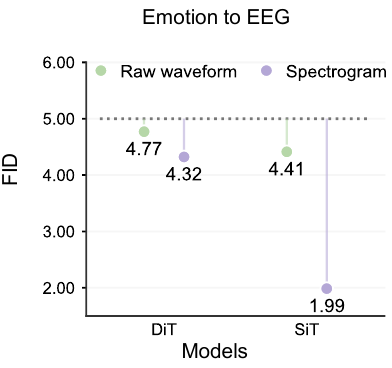}
    \captionof{figure}{\small{\textbf{High-frequency channel modeling approaches.}
    Spectrogram conditioning improves high-frequency channel generation.}}
    \label{fig:spectrogram_help}
\end{minipage}
\hfill
\begin{minipage}[t]{0.32\textwidth}
    \centering
    \includegraphics[width=\linewidth]{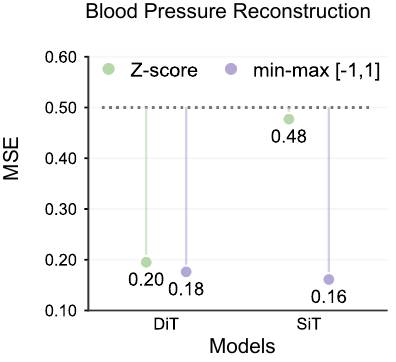}
    \captionof{figure}{\small{\textbf{Signal normalization strategies.}
    Range-based normalization outperforms z-score normalization for sensor-signal generation.}}
    \label{fig:normalization}
\end{minipage}
\hfill
\begin{minipage}[t]{0.31\textwidth}
    \centering
    \includegraphics[width=\linewidth]{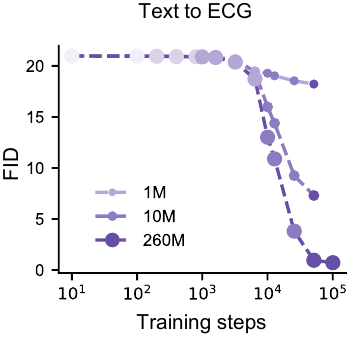}
    \captionof{figure}{\small{\textbf{Scaling behavior.}
    Increasing training steps and model capacity improves generation quality.}}
    \label{fig:scaling}
\end{minipage}

\end{figure*}

\textbf{Normalization choices matter for sensor generative modeling.}
We observe that z-score normalization can introduce undesirable generation bias, such as upward-shifted target signals in the VitalDB proxy-to-target translation task (see Appendix \ref{appendix:task_formulation}).
We therefore compare z-score normalization with range-based $\left[-1,1\right]$ min-max normalization on this setting.
As shown in Fig. \ref{fig:normalization}, range-based normalization consistently improves most metrics across models.
While z-score normalization is commonly effective for sensor representation learning \cite{narayanswamy2024scaling, shuai2026osf}, our results suggest that sensor generation may require different normalization choices, with fixed-range scaling providing a more stable target for generative training.

\finding{2}{Fixed-range normalization provides a more stable target for sensor generation.}

\subsection{Can sensor generative models scale?}

\textbf{Sensor generative models benefit from scaling compute and capacity.}
We further study whether increasing training computation and model capacity improves sensor-signal generation.
We report results on the text-to-ECG generation setting, where we evaluate SiT with different training steps and model sizes.
As shown in Fig.~\ref{fig:scaling}, generation quality improves consistently with longer training, as reflected by decreasing FID scores across all three models.
Increasing model size further improves signal fidelity, suggesting that sensor generative models benefit from both compute and capacity scaling.

\finding{3}{Sensor generative models scale with training compute and model capacity.}

%% file: sections/6_conclusion.tex
\section{Discussion}

\textbf{Limitations.}
\ours is a study of generative modeling for real-world sensor time series, and the generated signals should not be used for clinical diagnosis, treatment, or medical decision-making without further validation. Although \ours spans diverse domains, datasets, modalities, and generation settings, its coverage is still limited to the public datasets and task formulations included in this release; additional sensing environments, rare conditions, and deployment scenarios may reveal different model behaviors. Evaluating generated sensor signals also remains challenging: visually realistic waveforms may still miss important temporal, spectral, or physiological structure. Future work should extend \ours to broader sensor domains, stronger privacy and fairness analyses, clinically grounded utility metrics, and expert evaluation for high-stakes applications.

\textbf{Conclusion.}
We present \ours, a fully open, large-scale study of generative models for real-world sensor time series. Using \numtasksetting generation settings across \numdomain domains, \numdataset datasets, \nummodality signal modalities, and five representative generative model families, \ours enables controlled analysis of how task formulations, signal properties, model choices, and evaluation protocols shape generation quality. Our results show that flow-matching models serve as strong baselines across diverse settings, signal-aware designs improve generation in challenging regimes, and generated signals are useful beyond visual realism. Together, these findings provide practical lessons for developing and evaluating generative models for sensor time series, and establish \ours as an open foundation for future research on real-world sensor data generation.

%% file: sections/7_appendix.tex
\appendix

\section{Data Details}
\label{app:datasets}
% yuntian use claude to go through a pass
\subsection{Unified Data Infrastructure}
\input{tables/data_overview}
% we want to highlight the utility of our data pipline (unified data loading and preprocessing), so that we can hightlight our community contribution
In this section, we present the unified data preprocessing and preparation pipeline used in \ours. 
Rather than being a purely engineering component, this pipeline is a necessary foundation for our controlled study. 
It standardizes data collection, preprocessing, task construction, and data loading across heterogeneous sensor datasets, enabling consistent evaluation across domains, modalities, and task categories. 
The pipeline supports signals with diverse sampling frequencies, time spans, sequence lengths, and channel structures, allowing different generative model families to be trained and evaluated under the same framework. It enables the integration of isolated datasets and generation tasks, and thus enables revealing cross-setting findings in sensor-signal generation. 
Furthermore, our pipeline is also designed to be extensible, so that future datasets and task formulations can be incorporated with minimal modification. 
We believe this infrastructure can help support broader research on generative modeling for sensor time series.

\textbf{Preprocessing.}
The preprocessed data are stored in a shared HDF5 format.
Each signal is saved as a continuous array with shape (C,T), together with metadata such as sampling rate, channel names, subject identifier, recording/group identifier, and optional labels.
Task-specific resampling, windowing, and post-processing are performed at loading time, so the same stored files can support different task definitions without regeneration.
For datasets with heterogeneous sampling rates or modalities, signals from the same recording are aligned at the group level and stored as sibling entries in the shared HDF5 files. In the main experiments, each requested window loaded from the HDF5 files is normalized independently using per-channel min--max scaling to \(\left[-1, 1\right]\).

\textbf{Data organization and loading.}
We use a unified loading interface to construct task-specific examples from the shared HDF5 files. 
Instead of materializing all windows offline, the loader builds lightweight indices and reads only the requested windows during training to reduce I/O overhead. 
Specific periods of signals are sliced and loaded by time-stamp, resampled on the fly, concatenated across selected channels, and normalized when required. All tasks are represented by a common batch format, \texttt{UnifiedBatch}=\(\left(x,c_1,c_2,\mathrm{meta}\right)\), where $x$ is the target signal, $c_1$ contains sparse or global conditions such as text, labels, static covariates, or auxiliary channels, and $c_2$ contains dense temporal conditions such as past context or cross-channel signals. 
This abstraction covers all task categories in \ours, with light-weight model-specific adapters converting \texttt{UnifiedBatch} into the input format required by each generative model family.

% 1. make settings clear
% 2. highlight ours pipeline is a unified framework, reusable

\subsection{Data Usage Details}

Here we provide detailed statistics for the datasets used in our paper. 
Table~\ref{tab:data_stats_subjects} summarizes the number of subjects, domains, and modalities covered by each dataset, while Table~\ref{tab:data_freq_timespan_windows} reports the number of training and testing windows, sampling frequency, and time span of the preprocessed signals. As shown in the tables, our datasets span diverse real-world sensing domains and cover a broad range of modalities, resulting in signals with heterogeneous sampling frequencies and time spans. These diversities support our in-depth analysis upon generative modeling on sensor signals.

\clearpage
\section{Task Details}
\subsection{Task Formulation Details}
\label{appendix:task_formulation}

\paragraph{Notation.}
We formulate all benchmark settings as conditional sensor-signal generation. 
Let $\mathbf{x}\in\mathbb{R}^{C\times T}$ denote the target signal, where $C$ is the number of channels and $T$ is the sequence length. 
We use $c_1\in\mathbb{R}^{D}$ to denote the tokenized semantic condition, such as a text report, class label, or other non-signal metadata, and $c_2\in\mathbb{R}^{N\times D}$ to denote the tokenized signal-side condition derived from observed signals, such as historical context, source channels, temporal boundaries, low-resolution observations, noisy observations, or an observed signal to be edited. 
Here, $D$ is the embedding dimension and $N$ is the number of signal tokens.

Under this notation, each task defines a conditional distribution over target signals:
\[
    \hat{\mathbf{x}} \sim p_\theta\left(\mathbf{x}\mid \mathcal{C}\right),
    \qquad 
    \mathcal{C}\subseteq\{c_1,c_2\},
\]
where $\mathcal{C}$ denotes the set of available conditions for a specific task category. For example, semantic-to-signal tasks use only semantic conditions, while translation and temporal reconstruction tasks primarily use signal-side conditions. 

\paragraph{Semantic-to-Signal Generation.}
Semantic-to-signal generation synthesizes target signals from semantic conditions alone. Under our notation, the available condition set is $\mathcal{C}=\{c_1\}$, and the model generates
\[
    \hat{\mathbf{x}} \sim p_\theta\left(\mathbf{x}\mid c_1\right).
\]
The generated signal should be both realistic as a sensor signal and consistent with the semantic content specified by $c_1$. We therefore assess the realism and diversity of generated samples using distribution-level metrics, including FID, precision, and recall. This category includes two forms, depending on whether $c_1$ is represented by free-form text or a discrete label.

\paragraph{\interpextrap.}
Interpolation and extrapolation generate unobserved signal segments from observed temporal context. In all settings under this category, the model is given a signal-side condition $c_2$ that provides partial temporal observations, such as historical context or surrounding boundary segments. Some settings additionally include a compressed token as condition $c_1$, such as intervention variables or medication trajectories. The available condition set is therefore $\mathcal{C}=\{c_2\}$ or $\mathcal{C}=\{c_1,c_2\}$, and the model generates
\[
    \hat{\mathbf{x}} \sim p_\theta\left(\mathbf{x}\mid c_2\right)
    \quad \text{or} \quad
    \hat{\mathbf{x}} \sim p_\theta\left(\mathbf{x}\mid c_1,c_2\right).
\]
Here, forecasting-based tasks use past observations as $c_2$ and generate future segments, while temporal imputation uses surrounding observations as $c_2$ and generates the missing middle segment. We use sample-level evaluation metrics to measure the quality of generated signals.

\paragraph{\translation.}
Signal translation generates target modalities from observed signals. In this category, the model is given a signal-side condition $c_2$ corresponding to available channels or proxy measurements. Some tasks additionally include a compressed token as condition $c_1$, such as demographic covariates. The available condition set is therefore $\mathcal{C}=\{c_2\}$ or $\mathcal{C}=\{c_1,c_2\}$, and the model generates
\[
    \hat{\mathbf{x}} \sim p_\theta\left(\mathbf{x}\mid c_2\right)
    \quad \text{or} \quad
    \hat{\mathbf{x}} \sim p_\theta\left(\mathbf{x}\mid c_1,c_2\right).
\]
Here, $c_2$ and $\mathbf{x}$ correspond to different channels or modalities at the same timestamps. These tasks require models to recover target signals that are consistent with the observed source measurements. Therefore, we use sample-level metrics to evaluate reconstruction quality.

\paragraph{\signalediting.}

Given an observed signal $c_2$, the model generates a desired target signal $x$ while preserving non-targeted semantics whenever possible. The observed signal-side input may correspond to a low-resolution signal, a noisy signal, a partially observed signal, or a source signal to be edited. Some settings have additional signal-side conditions, such as correlated signals, which will also be treated as $c_2$. Semantic conditions $c_1$, including text instructions, target semantic conditions, or auxiliary covariates, are also provided in some tasks. Thus, $\mathcal{C}=\{c_2\}$ or $\mathcal{C}=\{c_1,c_2\}$, and
\[  
    \hat{\mathbf{x}} \sim p_\theta\left(\mathbf{x}\mid c_2\right)
    \quad \text{or} \quad
    \hat{\mathbf{x}} \sim p_\theta\left(\mathbf{x}\mid c_1,c_2\right).
\]

Note that observed signals can play different roles across tasks. For example, in glucose signal super-resolution, the low-resolution signal serves as a signal-side condition, whereas in ECG denoising, the noisy input signal is directly provided as the target to be restored by the generation model. The generated signal should follow the desired editing or restoration target, while preserving signal attributes that are not intended to change. Thus, we use sample-level metrics to assess generation fidelity.

\subsection{Task Setting Details}
\label{appendix:task details}
Here we provide details of our task settings organized by the datasets used in our study.

\subsubsection{MIMIC-IV ECG}
%logic here: category -- task
% then specific setting
\textbf{\semantictosignal{} -- Text-to-Signal Generation}
\begin{description}[
    itemsep=0pt,
    topsep=1pt,
    leftmargin=1em,
    labelwidth=0pt,
    labelsep=0pt,
    font=\normalfont
]
    \item[$\bullet$ Text-to-ECG Generation] \hfill \\
Inputs: Free-text ECG report $\mid$ Output: 12-lead ECG signal \\
Sequence Length: 1,000 $\mid$ Time Span: 10 sec $\mid$ Frequency: 100 Hz \\
Metric: FID, Precision, Recall
\end{description}

\textbf{\signalediting{} -- Semantic Signal Editing}
 \begin{description}[                   
      itemsep=0pt,                                                                                                                                    topsep=1pt,                        
      leftmargin=1em,
      labelwidth=0pt,
      labelsep=0pt,
      font=\normalfont                                                                                                                               
]
      \item[$\bullet$ ECG Editing] \hfill \\                                                                                 
Inputs: Source 12-lead ECG signal $x_a$ $\mid$ Source report $\text{text}_a$ $\mid$ Target report $\text{text}_b$ \\
Output: Edited 12-lead ECG signal $\hat{x}_b$ aligned with the target report \\                                                                      Sequence Length: 1{,}000 $\mid$ Time Span: 10 sec $\mid$ Frequency: 100 Hz \\                                                                
Metric: MSE, MAE, SMSE, PSNR, SSIM                                                                                  
\end{description}

\textbf{\signalediting{} -- Denoising}
\begin{description}[
      itemsep=0pt,
      topsep=1pt,
      leftmargin=1em,
      labelwidth=0pt,
      labelsep=0pt,
      font=\normalfont
  ]
      \item[$\bullet$ ECG Denoising] \hfill \\
Inputs: Noisy 12-lead ECG (\(x_{\text{noisy}} = x_{\text{clean}}\) + \(\sigma \cdot\varepsilon, \varepsilon \sim \mathcal{N}\left(0, I\right)\)) $\mid$ Free-text ECG report \\
Output: Denoised 12-lead ECG signal \\
Sequence Length: 1{,}000 $\mid$ Time Span: 10 sec $\mid$ Frequency: 100 Hz \\
Metric: MSE, MAE, SMSE, PSNR, SSIM
\end{description}

\subsubsection{PPG-DaLiA}
\textbf{\interpextrap{} -- Forecasting}
\begin{description}[
    itemsep=0pt,
    topsep=1pt,
    leftmargin=1em,
    labelwidth=0pt,
    labelsep=0pt,
    font=\normalfont
]
    \item[$\bullet$ Cardiac Signal Forecasting] \hfill \\
Inputs: historical HR, ECG, and BVP signals $\mid$ Output: future HR, ECG, and BVP signals \\
Sequence Length: 6,000 (4,000 context points; 2,000 target points) $\mid$ Time Span: 60 sec (40 sec context; 20 sec target) $\mid$ Frequency: 100 Hz \\
Metric: MSE, MAE, SMSE, PSNR, SSIM
\end{description}

\textbf{\translation{} -- Channel Translation}
\begin{description}[
    itemsep=0pt,
    topsep=1pt,
    leftmargin=1em,
    labelwidth=0pt,
    labelsep=0pt,
    font=\normalfont
]
    \item[$\bullet$ PPG-to-ECG Translation] \hfill \\
Inputs: PPG signal $\mid$ Output: ECG signal \\
Sequence Length: 6,000 $\mid$ Time Span: 60 sec $\mid$ Frequency: 100 Hz \\
Metric: MSE, MAE, SMSE, PSNR, SSIM
\end{description}

\subsubsection{SHHS}
\textbf{\interpextrap{} -- Temporal Imputation}
\begin{description}[
    itemsep=0pt,
    topsep=1pt,
    leftmargin=1em,
    labelwidth=0pt,
    labelsep=0pt,
    font=\normalfont
]
    \item[$\bullet$ Sleep Signal Imputation] \hfill \\
Inputs: left 10 sec and right 10 sec context from 2 EEG, 2 EOG, ECG, EMG $\mid$ Output: masked middle 10 sec of the same channels \\
Sequence Length: 1,920 (1,280 context points; 640 target points) $\mid$ Time Span: 30 sec (20 sec context; 10 sec target) $\mid$ Frequency: 64 Hz \\
Metric: MSE, MAE, SMSE, PSNR, SSIM
\end{description}

\subsubsection{PhyMER}
\textbf{\translation{} -- Channel Translation}
\begin{description}[
    itemsep=0pt,
    topsep=1pt,
    leftmargin=1em,
    labelwidth=0pt,
    labelsep=0pt,
    font=\normalfont
]
    \item[$\bullet$ Peripheral-to-EEG Translation] \hfill \\
Inputs: peripheral temperature, EDA, and BVP signals $\mid$ Output: 14-channel EEG signal \\
Sequence Length: 1,280 $\mid$ Time Span: 5 sec $\mid$ Frequency: 256 Hz \\
Metric: MSE, MAE, SMSE, PSNR, SSIM
\end{description}

\textbf{\semantictosignal{} -- Label-to-Signal Generation}
\begin{description}[
    itemsep=0pt,
    topsep=1pt,
    leftmargin=1em,
    labelwidth=0pt,
    labelsep=0pt,
    font=\normalfont
]
    \item[$\bullet$ Emotion-to-EEG Generation] \hfill \\
Inputs: emotion label $\mid$ Output: 14-channel EEG signal \\
Sequence Length: 1,280 $\mid$ Time Span: 5 sec $\mid$ Frequency: 256 Hz \\
Metric: FID, Precision, Recall
\end{description}

\subsubsection{CAPTURE-24}
\textbf{\semantictosignal{} -- Label-to-Signal Generation}
\begin{description}[
    itemsep=0pt,
    topsep=1pt,
    leftmargin=1em,
    labelwidth=0pt,
    labelsep=0pt,
    font=\normalfont
]
    \item[$\bullet$ Activity-to-IMU Generation] \hfill \\
Inputs: activity labels $\mid$ Output: 3-axis accelerometer signal \\
Sequence Length: 900 $\mid$ Time Span: 30 sec $\mid$ Frequency: 30 Hz \\
Metric: FID, Precision, Recall
\end{description}

\textbf{\signalediting{} -- Semantic Signal Editing}
 \begin{description}[                   
      itemsep=0pt,                                                                                                                                    topsep=1pt,                        
      leftmargin=1em,
      labelwidth=0pt,
      labelsep=0pt,
      font=\normalfont                                                                                                                               
]
      \item[$\bullet$ IMU Editing] \hfill \\                                                                                 
Inputs: Source 3-axis accelerometer signal $x_a$ $\mid$ Source label $\text{y}_a$ $\mid$ Target report $\text{y}_b$ \\
Output: Edited 3-axis accelerometer signal $\hat{x}_b$ aligned with the target report \\                                                                      Sequence Length: 900 $\mid$ Time Span: 30 sec $\mid$ Frequency: 30 Hz \\                                                                
Metric: MSE, MAE, SMSE, PSNR, SSIM                                                                                  
\end{description}

\subsubsection{VitalDB}
\textbf{\interpextrap{} -- Intervention-Conditioned Forecasting}
\begin{description}[
    itemsep=0pt,
    topsep=1pt,
    leftmargin=1em,
    labelwidth=0pt,
    labelsep=0pt,
    font=\normalfont
]
    \item[$\bullet$ Medication-Aware Forecasting] \hfill \\
Inputs: historical ECG, PPG, breathing, EEG, blood pressure, and medications $\mid$ Output: future ECG, PPG, breathing, EEG, and blood pressure \\
Sequence Length: 3,000 (2,000 context points; 1,000 target points) $\mid$ Time Span: 30 sec (20 sec context; 10 sec target) $\mid$ Frequency: 100 Hz \\
Metric: MSE, MAE, SMSE, PSNR, SSIM
\end{description}

\textbf{\translation{} -- Proxy-to-Target Reconstruction}
\begin{description}[
    itemsep=0pt,
    topsep=1pt,
    leftmargin=1em,
    labelwidth=0pt,
    labelsep=0pt,
    font=\normalfont
]
    \item[$\bullet$ Invasive Blood Pressure Reconstruction] \hfill \\
Inputs: PPG and non-invasive blood pressure $\mid$ Output: arterial blood pressure \\
Sequence Length: 1,500 $\mid$ Time Span: 30 sec $\mid$ Frequency: 50 Hz \\
Metric: MSE, MAE, SMSE, PSNR, SSIM
\end{description}

\subsubsection{Metabonet}
\textbf{\translation{} -- Channel Translation}
\begin{description}[
    itemsep=0pt,
    topsep=1pt,
    leftmargin=1em,
    labelwidth=0pt,
    labelsep=0pt,
    font=\normalfont
]
    \item[$\bullet$ Glucose Signal Translation] \hfill \\
Inputs: insulin, carbohydrate histories, and demographic covariates $\mid$ Output: continuous glucose monitoring (CGM) signal \\
Sequence Length: 2,016 $\mid$ Time Span: 7 days $\mid$ Frequency: 1 sample / 5 min \\
Metric: MSE, MAE, SMSE, PSNR, SSIM
\end{description}

\textbf{\signalediting{} -- Super-Resolution}
\begin{description}[
    itemsep=0pt,
    topsep=1pt,
    leftmargin=1em,
    labelwidth=0pt,
    labelsep=0pt,
    font=\normalfont
]
    \item[$\bullet$ Glucose Signal Super-Resolution] \hfill \\
Inputs: low-resolution 1-hour CGM signal, insulin, carbohydrate histories, and demographic covariates $\mid$ Output: high-resolution 5-min CGM signal \\
Sequence Length: 2,016 $\mid$ Time Span: 7 days $\mid$ Frequency: 1 sample / 5 min \\
Metric: MSE, MAE, SMSE, PSNR, SSIM
\end{description}

Overall, the benchmark covers 14 task settings across four generation families: \translation, \semantictosignal, \interpextrap, and \signalediting.

\clearpage
\section{Detailed Results}
\label{sec:appendix:detailed results}
\subsection{Detailed Results for Main Experiments}
In this section, we present the detailed setting-level results of the generation fidelity of representative generative model families across all task categories.
Overall, SiT provides the strongest overall baseline across the task suite, though no single family dominates every setting.
Specifically, in Table~\ref{tab:semantic_to_signal_distribution_level}, DiT and SiT achieve the strongest distribution-level fidelity in \semantictosignal{} settings, while SiT perform better in \interpextrap{} settings as shown in Table~\ref{tab:forecasting_sample_level} and Table~\ref{tab:imputation_sample_level}.
In \translation{} settings, Table~\ref{tab:translation_sample_level} and Table~\ref{tab:translation_sample_level_2} suggest that SiT remains competitive and performs best on PPG-to-ECG, invasive blood pressure reconstruction and glucose translation. 
And as marked in Table~\ref{tab:editing_sample_level} and Table~\ref{tab:cgm_super_resolution_sample_level}, SiT performs best on 2 of 4 \signalediting{} tasks. Beyond SiT's consistent advantage, the relative rankings of compared generative models vary across task categories. This suggests that modern generative paradigms do not naturally transfer uniformly to sensor-signal generation, and their inductive biases may align differently with sensor-specific requirements such as temporal dynamics, spectral structure, and conditional consistency. These findings have motivated us to analyze how these signal properties shape model preferences.

\subsection{Additional Quantitative Results}

In this section, we show the detailed results of our analysis studies.

\subsubsection{Synthetic Augmentation Enhances Downstream Sensor Signal Representations}
Building on the experimental settings detailed in the main text, we extend our evaluation to assess the practical downstream utility of synthetic sensor signals. Specifically, we investigate whether generated signals can effectively augment both self-supervised representation learning and supervised disease prediction. The motivation for this analysis is to determine whether generative models can transition from merely producing samples that look realistic to serving actual downstream clinical values.

\input{tables/appendix_translation}
\input{tables/appendix_forecast}
\input{tables/appendix_imputation_superres}
\input{tables/appendix_translation_2}
\input{tables/appendix_semantic2signal}
\input{tables/appendix_editing}
\input{tables/appendix_syn_data_help_sup}

For self-supervised learning, we evaluated downstream finetuning on PTB-XL using the encoders pre-trained on a mixture of real and synthetic data, as well as a baseline model pretrained only on the real data. We provided the detailed raw results in Table \ref{tab:synthetic_data_utility} as a supplementary to our main paper. We observe that a 50\% synthetic augmentation yields the optimal performance boost, effectively mitigating data scarcity as intended. Interestingly, aggressively scaling the synthetic augmentation to 100\% degrades performance relative to this peak, though it generally remains superior to using no synthetic data at all. %These results point to a few nuanced implications for data scaling in the sensor domain. 
The existence of an optimal mixing ratio is consistent with the intuition that while synthetic augmentation provides an immediate performance uplift, over-saturating the training pool likely forces the model to overfit to subtle generator artifacts or redundant, collapsed patterns. Consequently, this implies that future applications of generative biosignal models may benefit more from careful filtering, where only high-confidence or highly diverse synthetic samples are retained, rather than naive, brute-force dataset scaling.

Additionally, for the encoders pretrained with a mixture of real and synthetic data, and the baseline model pretrained with the real data, we test the zero-shot classification performance and report in Table~\ref{tab:synthetic_data_utility}. By adding synthetic data during SSL, as demonstrated in Table \ref{tab:synthetic_data_utility}, there are notable gains in the three zero-shot tasks, which again confirms the quality of the diverse and realistic signals generated by our model.

\input{tables/appendix_syn_data_help_ssl}

\subsubsection{Min-Max Normalization Mitigates Generative Vulnerability to Sensor Artifacts}
Sensor signals are inherently prone to noise originating from environmental factors, shifting subject conditions, and variable hardware quality. For example, the physical tightness of a respiratory band introduces substantial variance for identical breathing patterns, sudden motion during recordings causes extreme electrical spikes, and poor electrode contact frequently induces baseline wander.

While generative models excel at step-wise denoising (e.g., diffusion and flow-matching architectures), they are uniquely vulnerable to these extreme physiological artifacts. Because these models rely on mapping data to a bounded prior distribution, unbounded extreme values can destabilize the training objective. We hypothesized that bounding the signal amplitude using min-max normalization would provide a more stable learning target for these models compared to standard z-score normalization, which preserves highly extreme values and exposes the network to disproportionate gradient updates.

Following the analysis in the main paper, we provide the raw results here in Table \ref{tab:normalization_ablation_vitaldb}. The findings confirm our hypothesis: min-max normalization consistently matches or outperforms z-score normalization across all evaluated generative metrics.

These results carry an important practical implication for generative scaling in the sensor domain. They suggest that strictly bounding signal amplitudes is a more critical preprocessing step than preserving the global statistical variance of the raw dataset. Because z-score normalization preserves destructive, high-amplitude artifacts, bounded normalization strategies like min-max should be considered the default approach to ensure training stability and higher-fidelity generation in noise-prone physiological datasets.
\input{tables/appendix_signal_normalization}

\subsubsection{Generative models maintain long-sequence stability when signal-side context is available, but struggle in purely semantic-to-signal generation.}
We next examine whether introducing signal-side context can help mitigate quality degradation in long-sequence generation, since maintaining temporal consistency over extended horizons remains a primary challenge for many generative models. To this end, we evaluate paired short- and long-sequence settings across three distinct generation paradigms: semantic-to-signal generation, channel translation, and signal imputation. For comparable evaluation, we split each generated long signal into slices matching the short-sequence duration, compute metrics on each slice, and report the average.

Supplementary to the main paper, we provide the raw results in Table \ref{tab:long_sequence_evaluation} \& \ref{tab:capture24_long_sequence_distribution}. In case of \datasetcapturetwentyfour, generation quality severely degrades as sequence length increases, where the model must synthesize long-range accelerometry entirely from abstract activity type. Conversely, performance remains significantly more stable in imputation and translation settings. This contrast suggests that long-sequence generation is highly tractable when target-signal context is available, suggesting that the surrounding or parallel observed segments act as temporal anchors, effectively reducing the generative burden compared to unconstrained semantic-to-signal generation.
\input{tables/appendix_seq_better}
\input{tables/appendix_seq_worse}

\subsection{Failure Case Analysis}
\label{appendix:failure case}
\textbf{Modeling high-frequency signal is hard.}
We further examine scenarios where generative models fail to capture diverse signal semantics.
As shown in Table~\ref{tab:semantic_to_signal_distribution_level}, all model families struggle on the emotion-to-EEG generation task.
Although their precision scores remain relatively high, their recall scores are substantially lower.
This suggests that current generative models can synthesize realistic-looking EEG signals, but tend to cover only a limited subset of semantic and spectral patterns, failing to capture the full spectrum of EEG semantics.
This motivates us to study which design choices can improve generation quality in such challenging high-frequency signal settings.

\textbf{Frequency-domain metrics help evaluate high-frequency signal generation in more comprehensive aspects.}
As shown in Tables~\ref{tab:editing_sample_level} and~\ref{tab:translation_sample_level}, SSIM has limited discriminative power for high-frequency sensor signals, with scores concentrated in a narrow low-value range.
Yet qualitative examples indicate that the generated signals can still preserve realistic morphology.
This mismatch suggests that vision-derived structural metrics may not fully capture sensor-signal quality when semantics are encoded in high-frequency components.
We therefore introduce frequency-domain metrics for a more comprehensive evaluation of generated signals.

% \subsection{Are current metrics sufficient for evaluating signal generation quality?}

\input{tables/appendix_metric}

\textbf{Raw-waveform metrics can favor low-information predictions that are numerically close but dynamically unrealistic.}
To examine whether common reconstruction metrics fully reflect generation quality, we compare learned generators with deterministic historical-mean baselines on cardiac-signal forecasting, as an instantiation of the \interpextrap{} tasks categories.
The baseline summarizes the observed context by its channel-wise mean and repeats the resulting value across the target horizon.
As shown in Table~\ref{tab:ppgdalia_forecasting_multi_space}, although such predictions contain little temporal information, they can achieve competitive MSE and MAE by matching the average signal level.
In contrast, learned generative models produce signals with more realistic temporal dynamics and better distributional fidelity, even when their raw-waveform errors are not always lower.
This discrepancy shows that low time-domain error does not necessarily imply realistic sensor-signal generation.

\textbf{Feature-space metrics better align with signal realism.}
We further evaluate generated signals in a learned representation space by extracting embeddings for both generated and ground-truth signals and computing MSE and MAE between them.
As shown in Table~\ref{tab:ppgdalia_forecasting_multi_space}, compared with raw-waveform errors, the trends of feature-space errors are more consistent with FID scores, suggesting that representation-level evaluation may better captures morphology and realistic signal dynamics.
Together, these results show that sensor-generation quality cannot be assessed by a single metric family.

\finding{4}{Point-wise time-domain metrics alone are insufficient; robust signal-generation evaluation requires the combination of frequency, distribution, and representation level metrics.}

% unlock this for final version
\section{Detailed Qualitative Results}
\label{appendix:qualitative result}
In this section, we show additional qualitative examples of samples generated by our trained models. Here we show a few examples of the generated signals of representative generation settings. In general, we found ECG, blood pressure, glocuse signals, and breathing waveforms are easier to model, while high-frequency channels like EOG, EMG, and EEG channels are harder to model directly.

\begin{figure}
    \centering
    \includegraphics[width=1\linewidth]{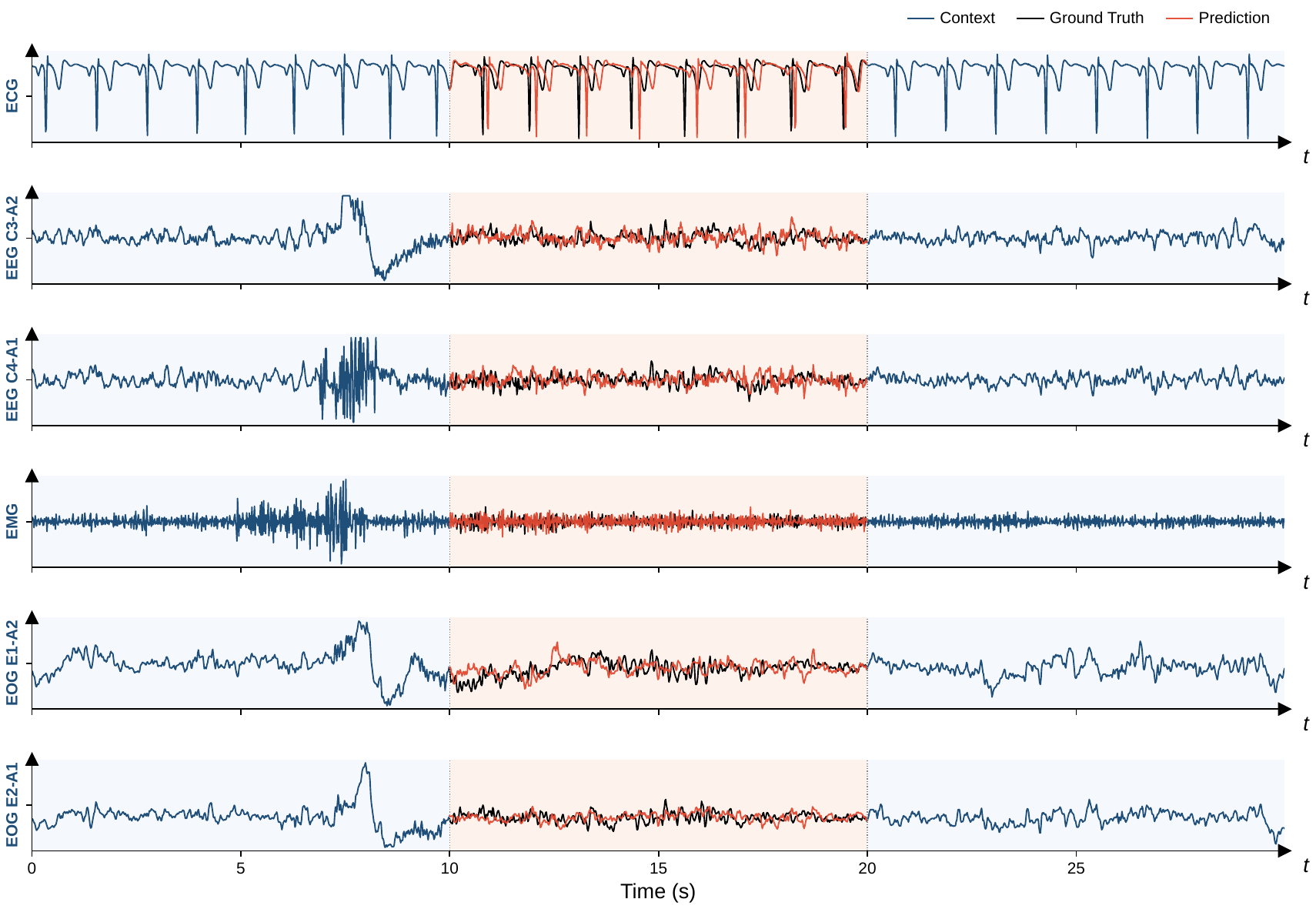}
    \caption{\textbf{Visualization of sleep signal imputation task.}}
    \label{fig:SHHS_imputation}
\end{figure}

\begin{figure}
    \centering
    \includegraphics[width=1\linewidth]{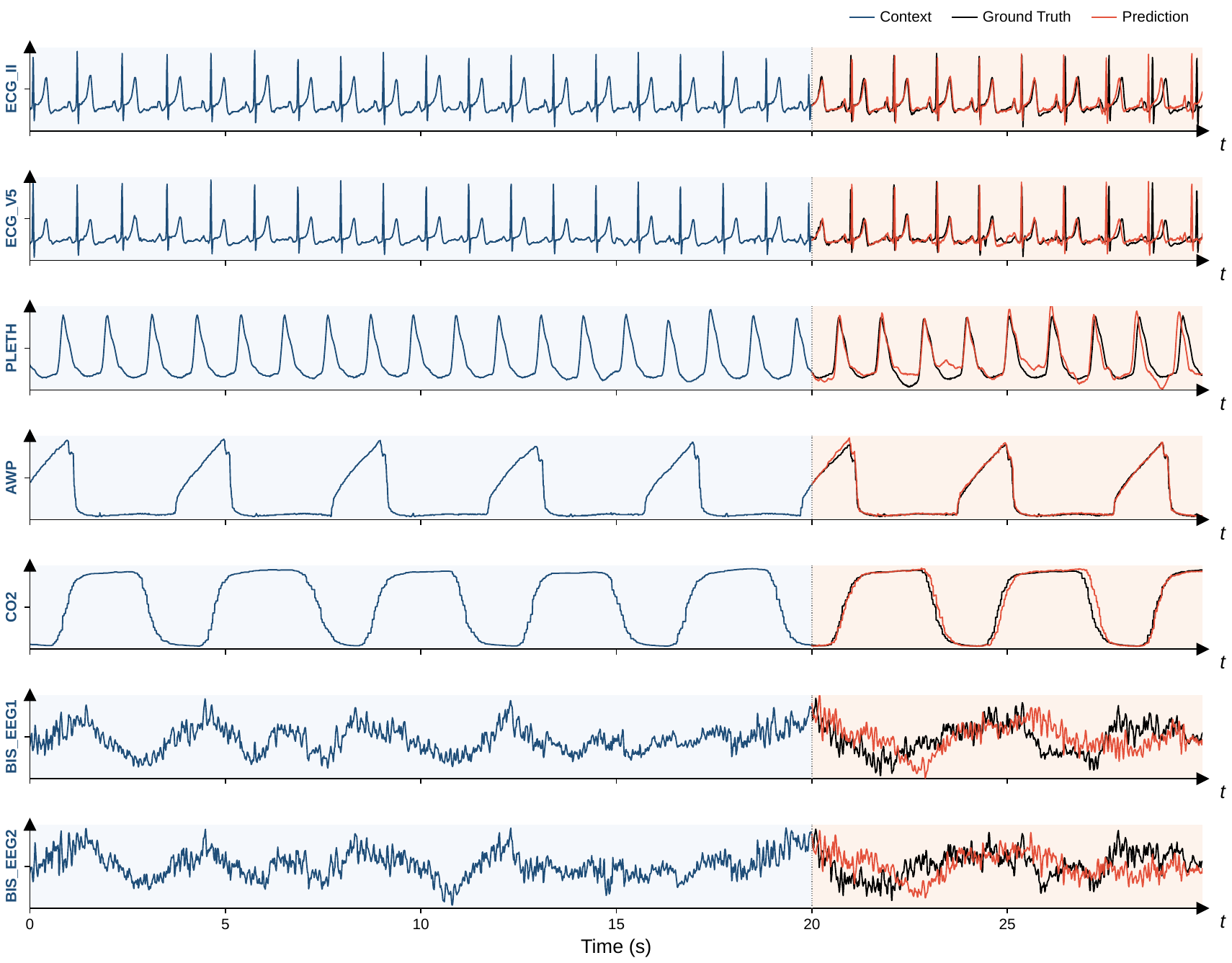}
    \caption{\textbf{Visualization of medication-aware forecasting task.}}
    \label{fig:VitalDB_forecasting}
\end{figure}

\begin{figure}
    \centering
    \includegraphics[width=1\linewidth]{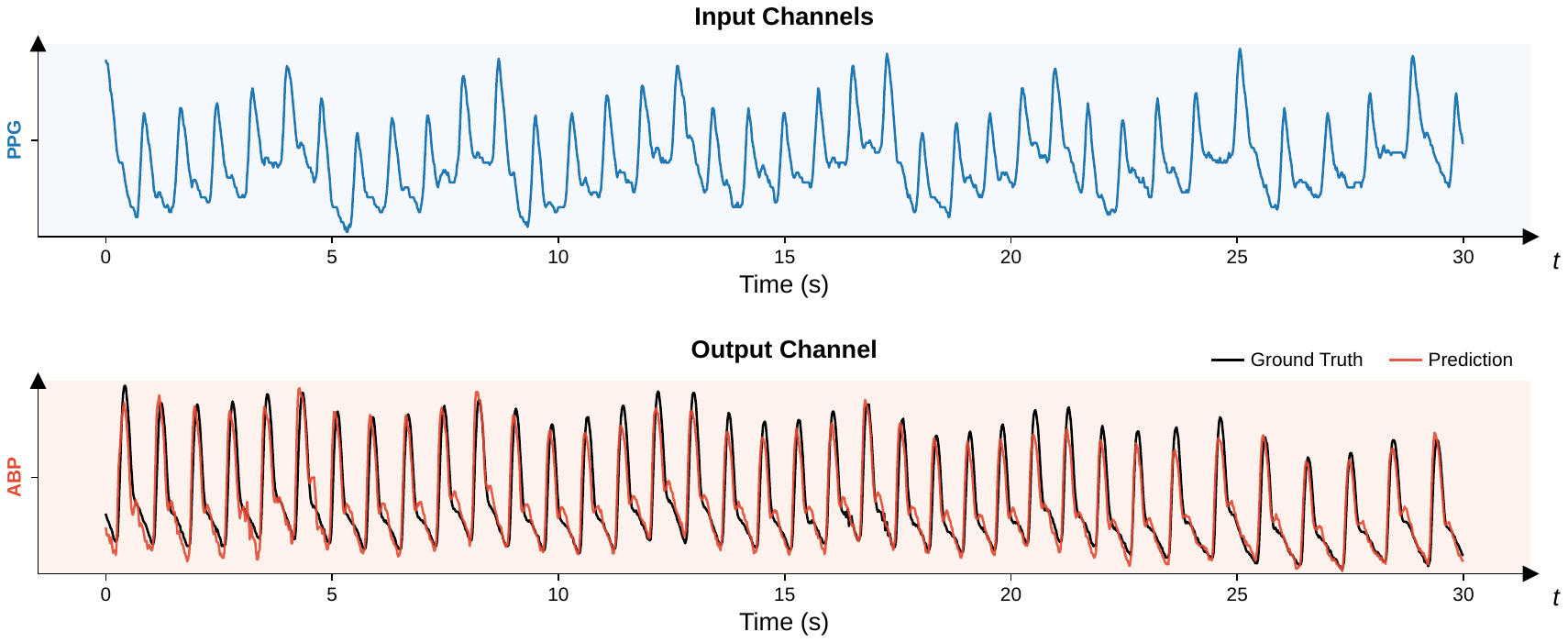}
    \caption{\textbf{Visualization of invasive blood pressure reconstruction task.}}
    \label{fig:VitalDB_BP_translate}
\end{figure}

\begin{figure}
    \centering
    \includegraphics[width=1\linewidth]{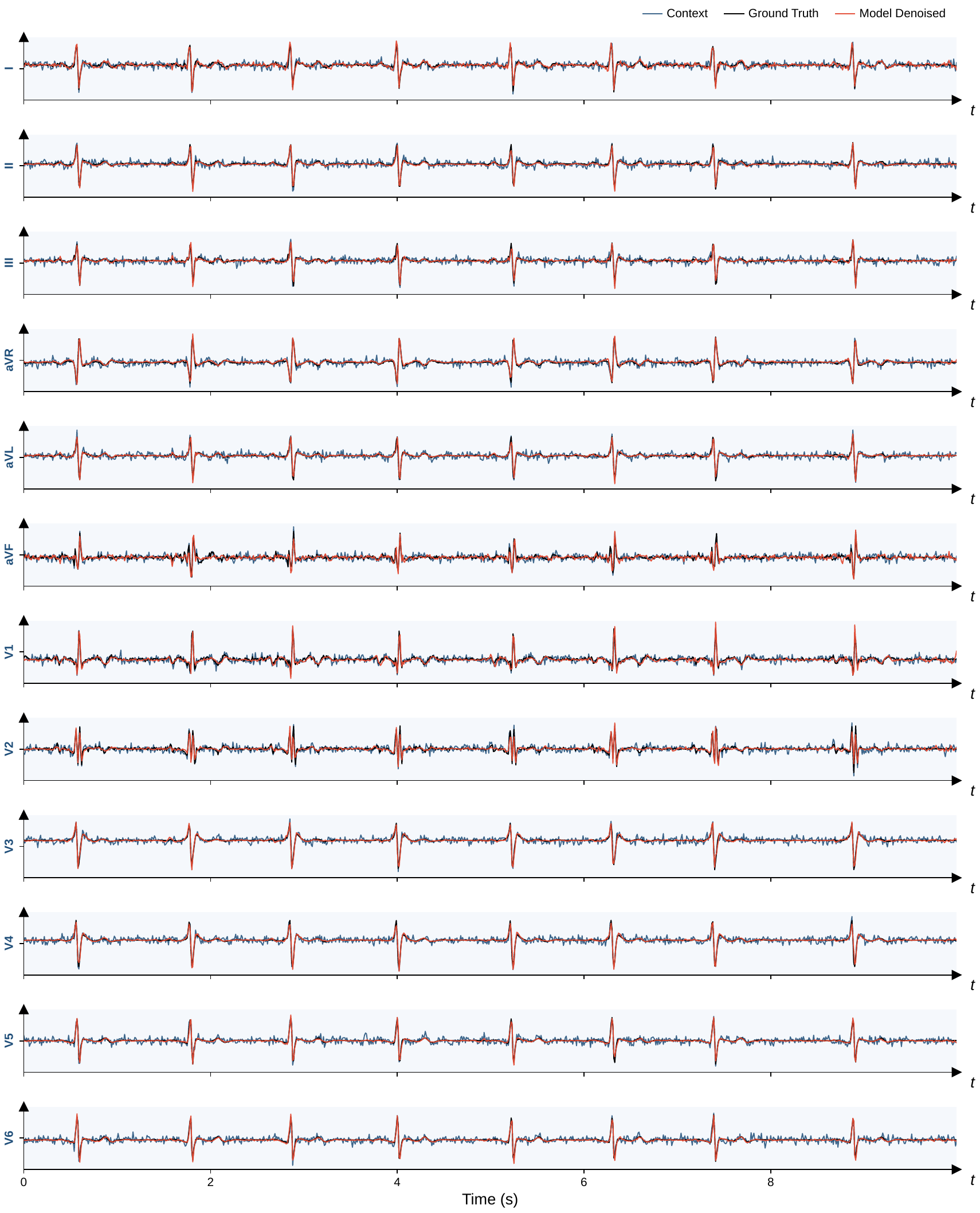}
    \caption{\textbf{Visualization of ECG denoising task.}}
    \label{fig:ecg denoising}
\end{figure}

\begin{figure}
    \centering
    \includegraphics[
        width=1\linewidth,
        trim=0 0 0 1.5cm,
        clip
    ]{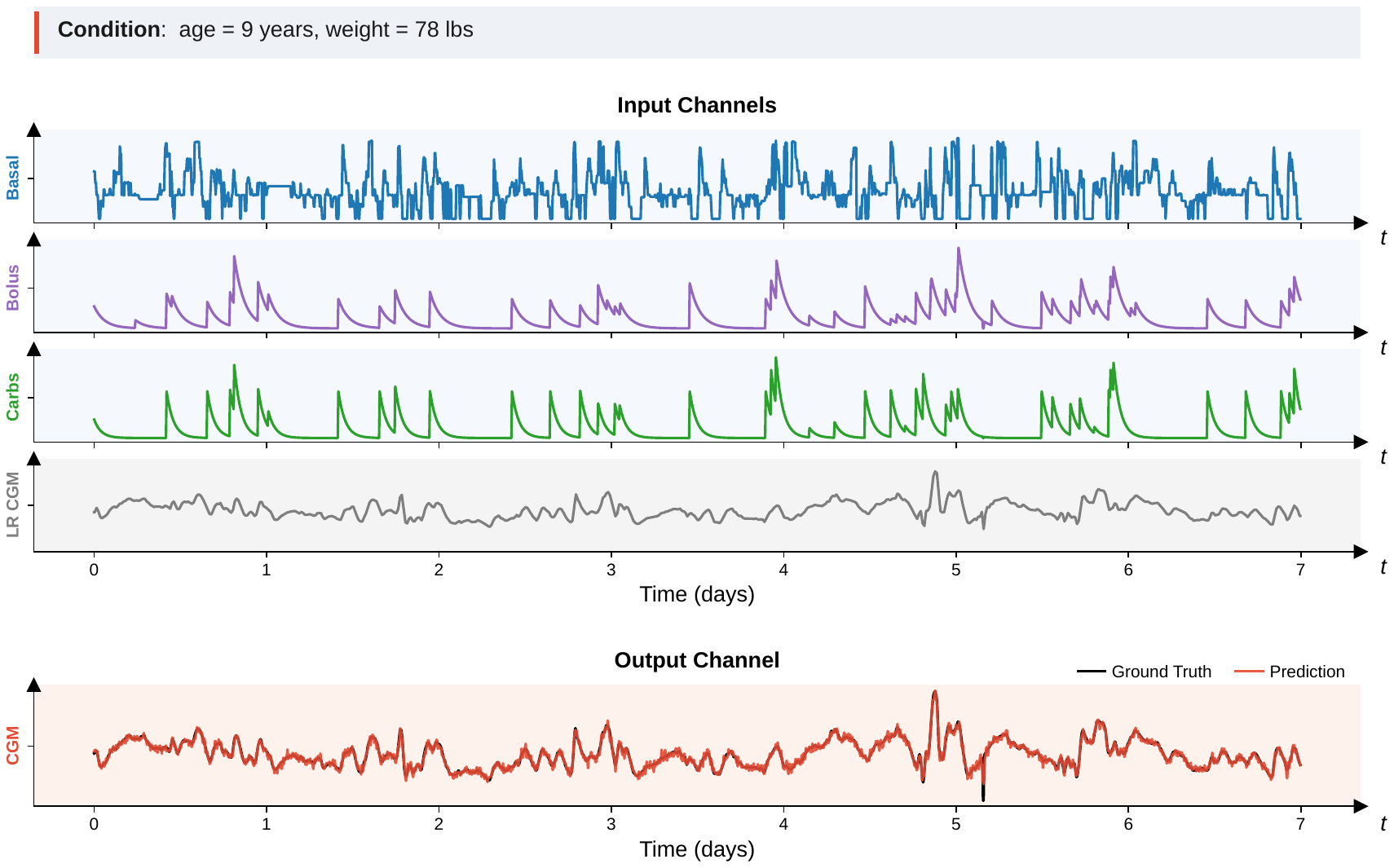}
    \caption{\textbf{Visualization of glucose signal super-resolution task.}}
    \label{fig:viz_glucose_superres}
\end{figure}
\begin{figure}
    \centering
    \includegraphics[width=1\linewidth]{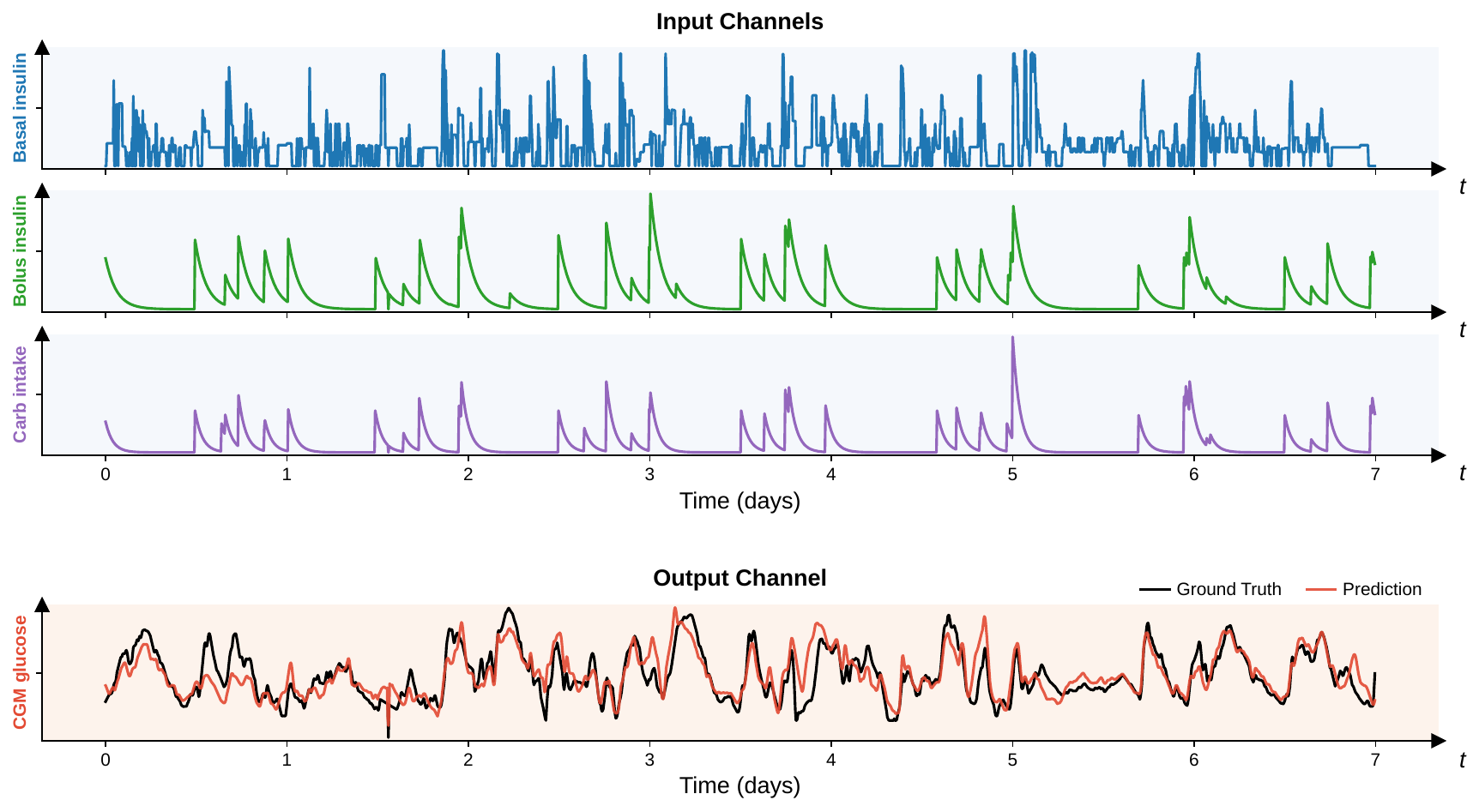}
    \caption{\textbf{Visualization of glucose signal translation task.}}
    \label{fig:cgm translation}
\end{figure}
\begin{figure}
    \centering
    \includegraphics[width=.95\linewidth]{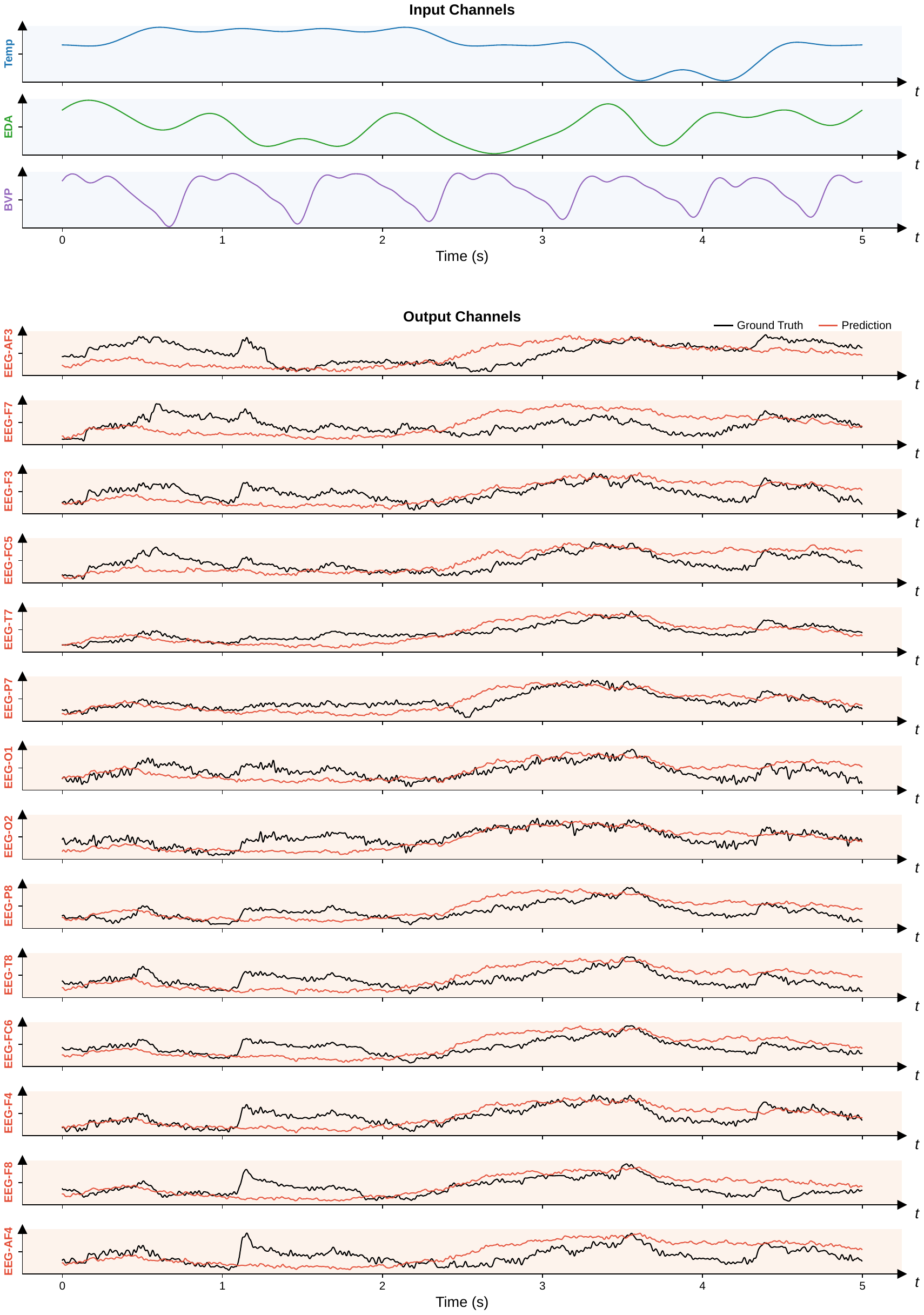}
    \caption{\textbf{Visualization of peripheral to Brain signal translation task.}}
    \label{fig:peripheral_to_brain}
\end{figure}

\clearpage
\section{Implementation Details}

% main setting
\label{appendix:label detail}
\subsection{Training Settings}
\label{appendix:training_settings}
\paragraph{Shared training protocol.}
% For each task setting, all model families use the same processed data splits and task-specific conditioning information, and all generative backbones are trained from scratch.
The main challenge in \ours is that different generative families have different native architectures, losses, samplers, and conditioning mechanisms. Appendix~\ref{appendix:label detail} therefore defines the shared experimental protocol used to make comparisons attributable to model behavior. For the main experiments, since we aim to understand how each modeling paradigm behaves in different sensor generation tasks, we didn't apply complex training and optimization techniques for each method, and avoided hyper-parameter searching if training curves are healthy and the generated samples are in reasonable morphologies. Therefore, we tried our best to align the training and evaluation settings of each method, unless otherwise specified, models are trained for up to \(50\mathrm{K}\) steps or until convergence with a per-rank batch size of \(64\), corresponding to a global batch size of \(256\) on \(4\times\) GH200 GPUs. Since the official implementations differ in architecture design and default configurations, we tried our best to do parameter matching across different architectural families. Table~\ref{tab:model_capacity_settings} summarizes the model-specific parameter-scale, learning-rate, EMA, and warmup configurations used by each model family. Table~\ref{tab:task_pretrain_settings} summarizes the training-side structure for the main benchmark tasks.

\paragraph{Detailed settings in analysis studies.}
For the analysis studies, we isolate one design factor at a time while keeping the remaining setup fixed whenever possible. For demographic encoding, we compare no demographic conditioning, raw demographic covariates,
and z-score normalized demographic covariates on the CGM translation task. 
For long-sequence
generation, we construct paired short- and long-sequence variants that differ only in generated
window length. We evaluate SHHS sleep imputation with \(1{,}920\) vs. \(19{,}200\) time steps,
CAPTURE-24 activity-to-IMU generation with \(900\) vs. \(9{,}000\) time steps, and PPG-to-ECG
translation with \(1{,}000\) vs. \(6{,}000\) time steps. To ensure comparable metrics, long generated
signals are sliced into segments matching the corresponding short-sequence duration, and metrics
are averaged across slices. For time-frequency modeling, we compare raw-waveform training with
STFT-magnitude spectrogram conditioning under the same emotion-to-EEG setting. For scaling
analysis, we vary training steps and model size while keeping the data, conditioning, sampling, and
evaluation protocol fixed. For signal normalization, we compare z-score normalization with
range-based min--max normalization to \([-1,1]\) on the same invasive blood pressure translation task setup. For
synthetic-data utility, we evaluate two uses of generated ECGs: real-plus-synthetic ECG--text
pretraining for representation transfer, and real-plus-synthetic supervised classifier training, each
compared against its real-only counterpart under the same downstream protocol.

\subsection{Evaluation Details}
\label{appendix:evaluation_settings}

In this section, we provide the evaluation protocol used in our study. 
Since \ours spans heterogeneous generation settings, we select metrics according to the structure of each task rather than enforcing a single metric family across all tasks. 
Semantic-to-signal tasks are evaluated with distribution-level feature-space metrics, paired generation tasks are evaluated with sample-level reconstruction metrics, and synthetic-data studies are evaluated through downstream utility.

\paragraph{Sampling protocol.}
Unless specified, for each model checkpoint in each task setting, we generate \(N_g=1024\) samples conditioned on held-out test examples. 
Unless otherwise specified, the main experiment uses either the checkpoint at the maximum training budget or the converged checkpoint within training budget. Sampling follows the default sampler provided by the official implementation of each model family.

\paragraph{Distribution-level metrics.}
For semantic-to-signal generation, multiple plausible signals may correspond to the same semantic condition. 
We therefore compare generated samples against the held-out real distribution using FID, Precision, and Recall in a learned signal-feature space. We use MIRA~\cite{li2025mira} to extract embedding and support the calculation of these metrics. 
We pick these metrics because they measure distributional alignment, sample fidelity, and distributional coverage~\cite{lai2025diffusets}.

% The feature encoders used for these metrics are summarized in Table~\ref{tab:eval_feature_encoders}.

\paragraph{Paired reconstruction metrics.}
For tasks with a unique paired target, we report MSE, MAE, SMSE, PSNR, and SSIM~\cite{wang2004image}. 
MSE and MAE quantify point-wise reconstruction error, while PSNR provides a logarithmic rescaling of reconstruction error relative to the signal range. SSIM measures window-based structural similarity, and SMSE captures spectral mismatch in log-power-spectral-density space, which is particularly informative for signals with substantial high-frequency content. Specifically, SMSE is computed as: 
% \begin{equation}
% \mathrm{SMSE}
% =
% \frac{1}{N}
% \sum_{n=1}^{N}
% \left\|
% \log\!\left(1 + \mathrm{PSD}\left(\hat{x}_{n}\right)\right)
% -
% \log\!\left(1 + \mathrm{PSD}\left(x_{n}\right)\right)
% \right\|_{2}^{2},
% \label{eq:smse}
% \end{equation}

\begin{equation}
\mathrm{SMSE}
=
\frac{1}{N C F}
\sum_{n=1}^{N}
\sum_{c=1}^{C}
\sum_{f=1}^{F}
\left(
\log\!\left(1+\mathrm{PSD}(\hat{x}_{n})[c,f]\right)
-
\log\!\left(1+\mathrm{PSD}(x_{n})[c,f]\right)
\right)^2 ,
\label{eq:smse}
\end{equation}
% where \(\hat{x}_{n}\) and \(x_n\) denote the generated and ground-truth signals for the \(n\)-th sample of length $T$, and $\mathrm{PSD}(x)[f] = \tfrac{1}{T}|\mathcal{F}_x[f]|^2$ denotes the one-sided power spectral density estimated from the squared FFT magnitude. Concretely, for a length-$T$ signal, $\mathrm{PSD}(x)[f] = \tfrac{1}{T}\,|\mathcal{F}_x[f]|^2$ where $\mathcal{F}_x = \mathrm{rfft}(x)$, and interior frequency bins ($0 < f < f_{\text{Nyquist}}$) are multiplied by 2 so that the one-sided spectrum preserves the total signal energy.

where \(\hat{x}_{n}\) and \(x_n\) denote the generated and ground-truth signals for the \(n\)-th sample, \(C\) is the number of channels, and \(F\) is the number of one-sided frequency bins. 

Paired metrics are computed in the normalized signal space used during training, typically with per-window min--max normalization to $[-1,1]$. Please check Appendix~\ref{app:datasets} for per-task normalization details.

\paragraph{Evaluation settings for analysis studies.}
For controlled analyses, we reuse the main sampling and evaluation protocol whenever possible, changing only the training variant, representation, or evaluation view required by the analysis.

\input{tables/appendix_presetting_modelspec}

\input{tables/appendix_presetting_tasks}

% \input{tables/appendix_presetting_modelvariant.tex}

% \input{tables/appendix_evalsetting_overview}   

% \input{tables/appendix_eval_syndata}

% \input{tables/appendix_eval_ftsetting}

% \input{tables/appendix_eval_analysisstudy}

% table format to present settings

\section{Model Implementation Details}
\label{appendix:model_detail}

We adapt representative families of generative models to time-series physiological signals using a unified implementation interface. 
Here we describe how each family is instantiated for multi-channel sensor time series. 
Across all models, the input signal is represented as \(x\in\mathbb{R}^{B\times C\times T}\), where \(B\), \(C\), and \(T\) denote the batch size, number of channels, and number of time steps, respectively.

\paragraph{Patchification.}
For transformer-based models, we replace 2-D image patchification with non-overlapping 1-D signal patchification \cite{narayanswamy2024scaling, yang2023simper}. 
Given a patch length \(P\), DiT, SiT, Imagen, MAR, and FractalGen convert the signal into a token sequence \(z\in\mathbb{R}^{B\times N\times D}\), where \(N=T/P\) and \(D\) is the hidden dimension. 
For DiT, SiT, and Imagen, this is implemented with a 1-D convolutional patch embedder.
The generated tokens are then projected back to the waveform domain through an unpatchification head. 
% TarFlow uses a different tokenization because invertible flows must preserve all input entries: it reshapes the signal into non-overlapping patches,
% \begin{equation}
% x\in\mathbb{R}^{B\times C\times T}
% \quad\mapsto\quad
% z\in\mathbb{R}^{B\times N\times CP}.
% \label{eq:tarflow_patch}
% \end{equation}

\paragraph{Conditioning approach.}
We have used sample-level semantic conditions, such as text reports, labels, and metadata, denoted as \(c_1\). And also dense signal-side conditions, such as historical signals, source channels, low-resolution references, temporal boundaries, or noisy observations, which are noted as \(c_2\). 
Specifically, DiT, SiT, and Imagen use sample-level modulation for \(c_1\) and cross-attention for \(c_2\); MAR uses \(c_1\) buffer tokens and decoder cross-attention for \(c_2\); FractalGen injects conditions across hierarchical levels; and TarFlow uses token-additive sparse conditioning for \(c_1\) with cross-attention for \(c_2\).
The codebase will be fully open after acceptance, and all implementation details can be found there.

% remember to illustrate how we did the c1, c2 conditioning

% how to do the adaptation from vision to signal

\subsection{Diffusion Models}

\paragraph{DiT.}
DiT~\cite{peebles2023scalable} is a representative transformer-based diffusion architecture that replaces the conventional U-Net denoiser with a scalable transformer backbone. Based on official implementation, we adapt DiT to 1-D physiological signals by replacing the 2-D image patch embedder with a 1-D patch embedder. 
The model is trained with the standard diffusion noise-prediction objective:
\begin{equation}
\mathcal{L}_{\mathrm{DiT}}
=
\mathbb{E}_{x_0,t,\epsilon}
\left[
\left\|
\epsilon
-
\epsilon_\theta\left(x_t,t,c_1,c_2\right)
\right\|_2^2
\right],
\label{eq:dit_loss}
\end{equation}
where $x_0$ denotes the data sample and \(x_t\) denotes the noisy signal at diffusion timestep \(t\). We follow the official implementation to inject the timestep embedding and sparse condition \(c_1\) through adaLN-zero modulation. We inject the dense condition \(c_2\) through cross-attention. 
The output tokens are projected back to waveform patches and unpatchified into the generated signal.

\subsection{Flow Matching}
\paragraph{SiT.}
SiT~\cite{ma2024sit} is a representative transformer-based flow-matching model that shares the scalable transformer design of DiT but replaces the Markovian diffusion objective with velocity prediction. 
We adapt SiT to 1-D physiological signals using the same patchification, positional embedding, backbone, and conditioning interface as DiT. 
Given a data sample \(x_0\) and a Gaussian endpoint \(x_1\sim\mathcal{N}(0,I)\), we use a linear interpolation path:
\begin{equation}
x_t
=
(1-t)x_0 + t x_1,
\qquad
t\sim\mathrm{Uniform}(0,1).
\label{eq:sit_path}
\end{equation}
The target velocity under this path is \(x_1-x_0\), and the model is trained with the flow-matching objective:
\begin{equation}
\mathcal{L}_{\mathrm{SiT}}
=
\mathbb{E}_{x_0,x_1,t}
\left[
\left\|
v_\theta\left(x_t,t,c_1,c_2\right)
-
\left(x_1-x_0\right)
\right\|_2^2
\right].
\label{eq:sit_loss}
\end{equation}

\subsection{Autoregressive Models}

\paragraph{MAR.}
MAR~\cite{li2024autoregressive} is a masked autoregressive generative model for continuous-valued data, avoiding the need for vector quantization. 
We adapt its image-generation implementation to 1-D physiological signals by applying masked autoregressive modeling over 1-D signal tokens. 
During training, the signal is patchified into a token sequence, a subset of tokens is masked, and the decoder predicts contextual representations for the masked positions. 
Each masked token is generated by a lightweight continuous diffusion head conditioned on its decoder representation. We inject sparse condition \(c_1\) through prepended tokens and the global conditioning path, and inject dense condition \(c_2\) through decoder cross-attention. 
At inference time, MAR generates signals by iterative unmasking, repeatedly predicting and revealing subsets of masked tokens according to the sampling schedule.

\subsection{Normalizing Flows}

\paragraph{TarFlow.}
TarFlow~\cite{zhai2024normalizing} is a transformer-based normalizing flow that enables exact likelihood training through invertible transformations. 
We adapt its official implementation to 1-D physiological signals by reshaping the waveform into non-overlapping patches. 
Let \(z^{(0)}\in\mathbb{R}^{B\times N\times CP}\) denote the reshaped signal patches. 
TarFlow applies a stack of invertible transformer blocks that transform the data representation into a latent variable \(z\). 
With affine coupling, each block predicts scale and shift parameters \((a_\ell,b_\ell)\) and applies
\begin{equation}
z^{(\ell+1)}
=
\left(z^{(\ell)} - b_\ell\right)\odot \exp\left(-a_\ell\right),
\label{eq:tarflow_affine}
\end{equation}
up to the block-specific permutation.

Since the transformation is invertible with a tractable Jacobian determinant, TarFlow is trained by exact negative log-likelihood:
\begin{equation}
\mathcal{L}_{\mathrm{TarFlow}}
=
\frac{1}{2}\mathbb{E}\left[\|z\|_2^2\right]
-
\mathbb{E}\left[\log|\det J|\right],
\label{eq:tarflow_nll}
\end{equation}
where \(z\) is the final latent variable and \(J\) is the Jacobian of the data-to-latent transformation.

We inject sparse condition \(c_1\) as an additive sample-level token, and inject dense condition \(c_2\) through cross-attention inside the invertible blocks. 
For tasks such as super-resolution, low-resolution signals are used as fixed conditioning inputs rather than concatenated as additional channels.

\subsection{Hierarchical Modeling Methods}

\paragraph{FractalGen.}
FractalGen~\cite{li2025fractal} is a hierarchical coarse-to-fine generative model. 
We adapt its official implementation to 1-D physiological signals by representing each signal through multiple temporal levels and recursively refining coarse tokens into finer temporal structure. 
Let \(x^{(0)},x^{(1)},\ldots,x^{(K)}\) denote representations from coarse to fine levels. 
The hierarchical generation process can be factorized as
\begin{equation}
p_\theta(x)
=
\prod_{k=0}^{K-1}
p_{\theta_k}\!\left(x^{(k+1)} \mid x^{(k)}, c_1, c_2\right),
\label{eq:fractal_factorization}
\end{equation}
where parent-level representations provide context for finer-level generation.

The upper levels of FractalGen use MAR-style iterative masked token prediction over continuous token representations, while the innermost level uses an autoregressive discretized loss over per-channel amplitude bins. 
Sparse conditions are injected into the top-level generator and then influence finer levels through parent-level representations, while dense conditions are passed through level-specific projection pathways when enabled.
This design enables long-horizon signal generation by modeling coarse temporal structure and fine-grained waveform details in a coarse-to-fine hierarchy.

\paragraph{Imagen.}
Imagen~\cite{saharia2022photorealistic} is a cascaded diffusion model for coarse-to-fine hierarchical generation. 
We adapt its official implementation to 1-D sensor signals by replacing the image-domain modules with the same 1-D transformer backbone and convolutional patchifier used in our DiT and SiT implementations. 
In our implementation, the first stage generates a low-resolution signal, and the second stage refines it to the target resolution:
\begin{equation}
x_{\mathrm{LR}}
\sim
p_{\theta_1}(\cdot\mid c_1, c_2),
\qquad
x_{\mathrm{HR}}
\sim
p_{\theta_2}(\cdot\mid c_1, c_2,\mathrm{Up}(x_{\mathrm{LR}})).
\label{eq:imagen_cascade}
\end{equation}

Both stages are trained with diffusion-style noise-prediction objectives. 
For super-resolution, we train only the second-stage model, since the low-resolution signal is directly provided by the task. 
The model follows the DiT/SiT-style conditioning pathway for \(c_1\) and \(c_2\), with the low-resolution signal additionally provided as second-stage conditioning. 
In the main experiments, we use deterministic low-resolution conditioning.

% \subsection{Spectrogram-domain variants.}
% For high-frequency EEG settings, we also implement spectrogram-domain DiT/SiT variants. Raw EEG signals are transformed with a complex short-time Fourier transform (STFT). The real and imaginary components are interleaved along the channel dimension, producing a time-frequency representation that is modeled by a 2-D spectrogram variant of the transformer. Generated spectrograms are converted back to raw waveforms using inverse STFT. This branch is used to test whether explicit time-frequency modeling improves the generation of high-frequency physiological channels. It changes only the modeling domain, while preserving the same task split, conditioning information, and evaluation protocol.
\subsection{Spectrogram-domain variants.}
For high-frequency EEG settings, we utilize raw-domain architecture augmented with with STFT-magnitude spectrogram conditioning. The STFT magnitude is encoded as a dense signal-side condition, while the main generator directly predicts raw EEG waveforms. This variant evaluates whether spectrogram conditioning improves high-frequency physiological signal generation.

% \subsection{Naive Baselines}

\subsection{Naive Baselines}

We also include a deterministic naive baselines for forecasting tasks with observed signal conditions $c_2$. These baselines do not train any parameters. Instead, they generate predictions by applying mean pooling to the provided historical context. For each target channel, $c_2$ contains the observed historical values. We pool these values over time and repeat the resulting scalar across the prediction horizon, producing a constant future trajectory for that channel. We evaluate the deterministic prediction $\hat{\mathbf{x}}$ against the paired target signal $\mathbf{x}$ using the same sample-level metrics as the learned generators.

%% file: tables/data_overview.tex
\begin{table}[ht]
    \small
    \centering
    \renewcommand{\arraystretch}{1.2}
    \caption{\textbf{Overview of datasets used in \ours.} }
    \label{tab:data_stats_subjects}

    \begin{tabularx}{\textwidth}{l l >{\raggedright\arraybackslash}X r}
        \toprule[1.5pt]
        \textbf{Dataset}
        & \textbf{Domain}
        & \textbf{Modality}
        & \textbf{Patients} \\
        \midrule\midrule

        \datasetmimiciecg
        & \domained
        & electrocardiography (ECG)
        & 154{,}075 \\

        \datasetppgdalia
        & \domaindailylife
        & \begin{tabular}[t]{@{}l@{}}
            photoplethysmography (PPG) \\
            electrocardiography (ECG) \\
            heart rate (HR)
          \end{tabular}
        & 15 \\

        \datasetshhs
        & \domainlab
        & \begin{tabular}[t]{@{}l@{}}
            electrocardiography (ECG) \\
            electroencephalography (EEG) \\
            electromyography (EMG) \\
            electrooculography (EOG)
          \end{tabular}
        & 6{,}663 \\

        \datasetphymer
        & \domainlab
        & \begin{tabular}[t]{@{}l@{}}
            electroencephalography (EEG) \\
            photoplethysmography (PPG) \\
            electrodermal activity (EDA) \\
            skin temperature (TEMP)
          \end{tabular}
        & 30 \\

        \datasetcapturetwentyfour
        & \domaindailylife
        & \begin{tabular}[t]{@{}l@{}}
            inertial measurement unit (IMU)
          \end{tabular}
        & 151 \\

        \datasetvitaldb
        & \domainoperationroom
        & \begin{tabular}[t]{@{}l@{}}
            electrocardiology (ECG) \\
            photoplethysmography (PPG) \\
            arterial blood pressure (ABP) \\
            non-invasive blood pressure (NIBP) \\
            airway pressure waveform (AWP) \\
            end-tidal capnography (Airflow) \\
            electroencephalography (EEG) \\
            medication volume
          \end{tabular}
        & 3{,}073 \\

        \datasetmetabonet
        & \domaindailylife
        & \begin{tabular}[t]{@{}l@{}}
            glucose monitoring (CGM) \\
            insulin delivery - basal  \\
            insulin delivery - bolus  \\
            carbohydrate intake 
          \end{tabular}
        & 1{,}211 \\

        \bottomrule[1.5pt]
    \end{tabularx}
\end{table}

\begin{table}[ht]
    \small
    \centering
    \renewcommand{\arraystretch}{1.2}
    \caption{\textbf{Data configurations used in \ours.} The segmented windows in each generation setting.}
    \label{tab:data_freq_timespan_windows}

    \begin{tabularx}{\textwidth}{l >{\raggedright\arraybackslash}X >{\raggedright\arraybackslash}X r r}
        \toprule[1.5pt]
        \textbf{Dataset}
        & \textbf{Frequency}
        & \textbf{Timespan}
        & \textbf{Train windows}
        & \textbf{Test windows} \\
        \midrule\midrule

        \datasetmimiciecg %emergency department
        & 100~Hz 
        & 10~s 
        & 533.3k
        & 227.3k \\

        \datasetppgdalia % daily life
        & 100~Hz
        & 1 min 
        & 1.4k
        & 0.7k \\

        \datasetshhs %lab study
        & 64~Hz
        & 30~s 
        & 5{,}550.9K
        & 1{,}590.4K \\

        \datasetphymer %lab study
        & 256~Hz
        & 5~s 
        & 12.2k
        & 3.1k \\

        \datasetcapturetwentyfour % daily life
        & 30~Hz
        & 30~s 
        & 245.1k
        & 62.1k \\

        \datasetvitaldb % operation room
        & 50/100~Hz
        & 30~s
        & 826.4k
        & 204.7k \\

        \datasetmetabonet % daily life
        & 12 per Hour  
        & 7~days 
        & 360.3k
        & 63.8k \\

        \bottomrule[1.5pt]
    \end{tabularx}
\end{table}

%% file: tables/appendix_translation.tex
\begin{table*}[!t]
  \centering
  \caption{\textbf{Performance comparison on \translation.} \sit\ performs better on PPG-to-ECG translation while \fractalgen\ is strong on Peripheral-to-EEG translation.}
  \label{tab:translation_sample_level}
  \vspace{-2mm}
  \begin{adjustbox}{width=\textwidth}
  \footnotesize
  \setlength{\tabcolsep}{3pt}
  \begin{tabular}{l *{5}{r} *{5}{r}}
    \toprule[1.5pt]
    \multirow{2.5}{*}{\textbf{Model}} &
    \multicolumn{5}{c}{\textbf{PPG-to-ECG Translation}} &
    \multicolumn{5}{c}{\textbf{Peripheral-to-EEG Translation}} \\
    \cmidrule(lr){2-6}\cmidrule(lr){7-11}
    & MSE$^\downarrow$ & MAE$^\downarrow$ & SMSE$^\downarrow$ & PSNR$^\uparrow$ & SSIM$^\uparrow$
    & MSE$^\downarrow$ & MAE$^\downarrow$ & SMSE$^\downarrow$ & PSNR$^\uparrow$ & SSIM$^\uparrow$ \\
    \midrule
    \midrule

    \dit
      & 0.257 & 0.370 & 0.043 & 12.37 & 0.207
      & 0.348 & 0.475 & 0.053 & 10.73 & 0.008 \\

    \sit
      & \textbf{0.244} & \textbf{0.363} & \textbf{0.041} & \textbf{12.52} & \textbf{0.219}
      & 0.352 & 0.479 & 0.053 & 10.68 & 0.009 \\

    \fractalgen
      & 0.381 & 0.472 & 0.173 & 10.26 & 0.022
      & \textbf{0.328} & \textbf{0.462} & 0.065 & \textbf{10.93} & 0.007 \\

    \tarflow
      & 0.281 & 0.398 & 0.052 & 11.83 & 0.076
      & 0.354 & 0.481 & \textbf{0.048} & 10.67 & 0.006 \\

    \mar
      & 0.267 & 0.378 & 0.049 & 12.17 & 0.178
      & 0.411 & 0.520 & 0.053 & 10.04 & \textbf{0.010} \\

    \bottomrule[1.5pt]
  \end{tabular}
  \end{adjustbox}
  \vspace{-3mm}
\end{table*}

%% file: tables/appendix_forecast.tex
\begin{table*}[!t]
  \centering
  \caption{\textbf{Forecasting and intervention-conditioned forecasting results.} \sit{} performs best on medication-aware forecasting and remains competitive with \mar{} on PPG-DaLIA forecasting. %We report two settings in the \interpextrap task category.
  }
  \label{tab:forecasting_sample_level}
  \vspace{-2mm}
  \begin{adjustbox}{width=\textwidth}
  \footnotesize
  \setlength{\tabcolsep}{3pt}
  \begin{tabular}{l *{5}{r} *{5}{r}}
    \toprule[1.5pt]
    \multirow{2.5}{*}{\textbf{Model}} &
    \multicolumn{5}{c}{\textbf{Medication-Aware Forecasting}} &
    \multicolumn{5}{c}{\textbf{Cardiac Signal Forecasting}} \\
    \cmidrule(lr){2-6}\cmidrule(lr){7-11}
    & MSE$^\downarrow$ & MAE$^\downarrow$ & SMSE$^\downarrow$ & PSNR$^\uparrow$ & SSIM$^\uparrow$
    & MSE$^\downarrow$ & MAE$^\downarrow$ & SMSE$^\downarrow$ & PSNR$^\uparrow$ & SSIM$^\uparrow$ \\
    \midrule
    \midrule

    \dit
      & 0.494 & 0.582 & 0.163 & 9.11 & 0.006
      & 0.224 & 0.372 & 0.046 & 13.20 & 0.185 \\

    \sit
      & \textbf{0.282} & \textbf{0.426} & 0.076 & \textbf{11.65} & 0.078
      & 0.203 & 0.345 & \textbf{0.039} & \textbf{13.68} & \textbf{0.231} \\

    \fractalgen
      & 0.532 & 0.509 & \textbf{0.068} & 9.02 & \textbf{0.133}
      & 0.206 & 0.344 & 0.049 & 13.21 & 0.124 \\

    \tarflow
      & 0.547 & 0.556 & 0.098 & 8.77 & 0.068
      & 0.223 & 0.370 & 0.052 & 13.18 & 0.131 \\

    \mar
      & 0.570 & 0.556 & 0.169 & 8.51 & 0.096
      & \textbf{0.198} & \textbf{0.340} & 0.043 & 13.52 & 0.175 \\

    \bottomrule[1.5pt]
  \end{tabular}
  \end{adjustbox}
  \vspace{-3mm}
\end{table*}

%% file: tables/appendix_imputation_superres.tex
\begin{table*}[!t]
  \centering
  \begin{minipage}[t]{0.49\textwidth}
  \centering
  \caption{\textbf{Performance comparison on \interpextrap.} \sit\ performs better on Sleep-Signal Imputation.}
  \label{tab:imputation_sample_level}
  \begin{adjustbox}{width=\linewidth}
  \footnotesize
  \setlength{\tabcolsep}{3pt}
  \begin{tabular}{l *{5}{r}}
    \toprule[1.5pt]
    \multirow{2.5}{*}{\textbf{Model}} &
    \multicolumn{5}{c}{\textbf{Sleep Signal Imputation}} \\
    \cmidrule(lr){2-6}
    & MSE$^\downarrow$ & MAE$^\downarrow$ & SMSE$^\downarrow$ & PSNR$^\uparrow$ & SSIM$^\uparrow$ \\
    \midrule
    \midrule

    \dit
      & 0.286 & 0.389 & 0.079 & 12.13 & 0.067 \\

    \sit
      & \textbf{0.270} & \textbf{0.381} & \textbf{0.077} & \textbf{12.25} & \textbf{0.077} \\

    \fractalgen
      & 0.296 & 0.385 & 0.092 & 12.18 & 0.071 \\

    \tarflow
      & 0.290 & 0.424 & 0.100 & 11.70 & 0.015 \\

    \mar
      & 0.286 & 0.398 & 0.091 & 12.01 & 0.077 \\

    \bottomrule[1.5pt]
  \end{tabular}
  \end{adjustbox}
  \end{minipage}
  \hfill
  \begin{minipage}[t]{0.49\textwidth}
  \centering
  \caption{\textbf{Performance comparison on \signalediting.} \imagen\ performs better on CGM super-resolution.}
  \label{tab:cgm_super_resolution_sample_level}
  \begin{adjustbox}{width=\linewidth}
  \footnotesize
  \setlength{\tabcolsep}{3pt}
  \begin{tabular}{l *{5}{r}}
    \toprule[1.5pt]
    \multirow{2.5}{*}{\textbf{Model}} &
    \multicolumn{5}{c}{\textbf{Glucose Signal Super-Resolution}} \\
    \cmidrule(lr){2-6}
    & MSE$^\downarrow$ & MAE$^\downarrow$ & SMSE$^\downarrow$ & PSNR$^\uparrow$ & SSIM$^\uparrow$ \\
    \midrule
    \midrule

    \dit
      & 0.059 & 0.151 & 0.032 & 22.05 & 0.339 \\

    \sit
      & 0.134 & 0.260 & 0.088 & 19.35 & 0.276 \\

    \fractalgen
      & 0.127 & 0.256 & 0.042 & 15.79 & 0.314 \\

    \tarflow
      & 0.034 & 0.144 & 0.008 & 21.55 & 0.370 \\

    \mar
      & 0.266 & 0.393 & 0.121 & 12.04 & 0.118 \\

    \imagen
      & \textbf{0.008} & \textbf{0.064} & \textbf{0.002} &\textbf{ 27.83 }& \textbf{0.699} \\

    \bottomrule[1.5pt]
  \end{tabular}
  \end{adjustbox}
  \end{minipage}
\end{table*}

%% file: tables/appendix_translation_2.tex
\begin{table*}[!t]
  \centering
  \caption{\textbf{Performance comparison on \translation.} \sit\ performs better on all these settings.}
  \label{tab:translation_sample_level_2}
  \vspace{-2mm}
  \begin{adjustbox}{width=\textwidth}
  \footnotesize
  \setlength{\tabcolsep}{3pt}
  \begin{tabular}{l *{5}{r} *{5}{r}}
    \toprule[1.5pt]
    \multirow{2.5}{*}{\textbf{Model}} &
    \multicolumn{5}{c}{\textbf{Glucose Signal Translation}} &
    \multicolumn{5}{c}{\textbf{Invasive Blood Pressure Reconstruction}} \\
    \cmidrule(lr){2-6}\cmidrule(lr){7-11}
    & MSE$^\downarrow$ & MAE$^\downarrow$ & SMSE$^\downarrow$ & PSNR$^\uparrow$ & SSIM$^\uparrow$
    & MSE$^\downarrow$ & MAE$^\downarrow$ & SMSE$^\downarrow$ & PSNR$^\uparrow$ & SSIM$^\uparrow$ \\
    \midrule
    \midrule

    \dit
      & 0.097 & 0.223 & 0.032 & 17.38 & 0.374
      & 0.176 & 0.283 & 0.045 & 13.56 & \textbf{0.445} \\

    \sit
      & \textbf{0.091} & \textbf{0.216} & \textbf{0.030} & \textbf{17.70} & \textbf{0.400}
      & \textbf{0.161} & \textbf{0.274} & \textbf{0.038} & \textbf{13.96} & 0.438 \\

    \fractalgen
      & 0.149 & 0.284 & 0.044 & 14.28 & 0.260
      & 0.571 & 0.598 & 0.114 & 8.46 & 0.113 \\

    \tarflow
      & 0.307 & 0.438 & 0.063 & 11.15 & 0.053
      & 0.584 & 0.611 & 0.226 & 8.53 & 0.106 \\

    \mar
      & 0.298 & 0.432 & 0.063 & 11.28 & 0.070
      & 0.571 & 0.601 & 0.218 & 8.45 & 0.105 \\

    %\imagen
     % & 0.215 & 0.359 & 0.055 & 13.05 & 0.149
     % & - & - & - & - & - \\

    \bottomrule[1.5pt]
  \end{tabular}
  \end{adjustbox}
  \vspace{-3mm}
\end{table*}

%% file: tables/appendix_semantic2signal.tex
\begin{table*}[!t]
  \centering
  \caption{\textbf{Performance comparison on \semantictosignal.} \sit\ achieves overall better performance on activity-to-IMU and emotion-to-EEG generation, and \imagen achieves better performance on text-to-ECG task setting.}
  \label{tab:semantic_to_signal_distribution_level}
  \vspace{-2mm}
  \begin{adjustbox}{width=\textwidth}
  \footnotesize
  \setlength{\tabcolsep}{4pt}
  \begin{tabular}{l *{3}{r} *{3}{r} *{3}{r}}
    \toprule[1.5pt]
    \multirow{2.5}{*}{\textbf{Model}} &
    \multicolumn{3}{c}{\textbf{Text-to-ECG}} &
    \multicolumn{3}{c}{\textbf{Activity-to-IMU}} &
    \multicolumn{3}{c}{\textbf{Emotion-to-EEG}} \\
    \cmidrule(lr){2-4}\cmidrule(lr){5-7}\cmidrule(lr){8-10}
    & FID$^\downarrow$ & Precision$^\uparrow$ & Recall$^\uparrow$
    & FID$^\downarrow$ & Precision$^\uparrow$ & Recall$^\uparrow$
    & FID$^\downarrow$ & Precision$^\uparrow$ & Recall$^\uparrow$ \\
    \midrule
    \midrule

    \dit
      & 1.07 & 0.982 & 0.373
      & 0.32 & 0.890 & 0.618
      & 4.77 & 0.788 & 0.098 \\

    \sit
      & 2.03 & \textbf{0.992} & 0.325
      & \textbf{0.25} & 0.893 & \textbf{0.683}
      & \textbf{4.41} & \textbf{0.861} & \textbf{0.232} \\

    \fractalgen
      & 11.89 & 0.963 & 0.059
      & 1.10 & 0.884 & 0.151
      & 8.83 & 0.049 & 0.008 \\

    \tarflow
      & 4.66 & 0.971 & 0.107
      & 0.79 & 0.857 & 0.225
      & 7.34 & 0.024 & 0.000 \\

    \mar
      & 14.48 & 0.702 & 0.001
      & 0.47 & \textbf{0.903} & 0.463
      & 8.33 & 0.637 & 0.011 \\

    \imagen
      & \textbf{1.07} & 0.976 & \textbf{0.478}
      & 0.54 & 0.830 & 0.550
      & 5.93 & 0.395 & 0.010 \\

    \bottomrule[1.5pt]
  \end{tabular}
  \end{adjustbox}
  \vspace{-3mm}
\end{table*}

%% file: tables/appendix_editing.tex
\begin{table}[!t]
  \centering
  \caption{\textbf{Performance comparison on training-free \signalediting.} \sit\ achieves the best performance on ECG editing and ECG denoising, and \dit is the strongest on IMU editing task.}
  \label{tab:editing_sample_level}
  \begin{adjustbox}{width=0.5\linewidth}
  \footnotesize
  \setlength{\tabcolsep}{3pt}
  \begin{tabular}{l *{5}{r}}
    \toprule[1.5pt]
    \multirow{2.5}{*}{\textbf{Model}} &
    \multicolumn{5}{c}{\textbf{Sample Level}} \\
    \cmidrule(lr){2-6}
    & MSE$^\downarrow$ & MAE$^\downarrow$ & SMSE$^\downarrow$ & PSNR$^\uparrow$ & SSIM$^\uparrow$ \\
    \midrule
    \midrule

    \multicolumn{6}{l}{\textit{ECG Editing}} \\
    \dit        & 0.138 & 0.253 & 0.045 & 14.89 & 0.272 \\
    \sit        & \textbf{0.129} & \textbf{0.236} & \textbf{0.043} & \textbf{15.16} & \textbf{0.287} \\
    \midrule

    \multicolumn{6}{l}{\textit{ECG Denoising}} \\
    \dit        & 0.034 & 0.099 & 0.012 & 23.74 & 0.624 \\
    \sit        & \textbf{0.006} & \textbf{0.051} & \textbf{0.002} & \textbf{29.05} & \textbf{0.739} \\
    \midrule

    \multicolumn{6}{l}{\textit{IMU Editing}} \\
    \dit        & \textbf{0.308} & \textbf{0.447} & \textbf{0.074} & \textbf{11.58} & \textbf{0.001} \\
    \sit        & 0.329 & 0.463 & 0.085 & 11.22 & 0.000 \\

    \bottomrule[1.5pt]
  \end{tabular}
  \end{adjustbox}
\end{table}

%% file: tables/appendix_syn_data_help_sup.tex
\begin{table}[!t]
  \centering
  \caption{\textbf{Downstream utility of generated signals.}
    With ECGs synthesized by a text-to-ECG generator, we evaluate their utility supervised learning, and found moderate synthetic augmentation improves disease classification performance, whereas excessive synthetic data degrades it.}
  \label{tab:synthetic_data_augmentation}
  \begin{adjustbox}{width=0.65\linewidth}
  \footnotesize
  \setlength{\tabcolsep}{3pt}
  \begin{tabular}{l l *{6}{r}}
    \toprule[1.5pt]
    \multirow{2.5}{*}{\textbf{Training Data}} &
    \multirow{2.5}{*}{\textbf{Ratio}} &
    \multicolumn{2}{c}{\textbf{MI}} &
    \multicolumn{2}{c}{\textbf{STTC}} &
    \multicolumn{2}{c}{\textbf{CD}} \\
    \cmidrule(lr){3-4}\cmidrule(lr){5-6}\cmidrule(lr){7-8}
    & & AUC$^\uparrow$ & AUPRC$^\uparrow$
      & AUC$^\uparrow$ & AUPRC$^\uparrow$
      & AUC$^\uparrow$ & AUPRC$^\uparrow$ \\
    \midrule
    \midrule

    \textsc{Real}
      & -
      & 90.86 & 70.62
      & 92.17 & 73.37
      & 90.14 & 74.64 \\

    \midrule

    \multirow{3}{*}{\textsc{Real + Syn}}
      & 10\%
      & 91.31 & \textbf{74.25}
      & \textbf{92.95} & \textbf{74.33}
      & 91.56 & 76.45 \\
      & 50\%
      & \textbf{91.51} & 73.11
      & 92.83 & 72.29
      & \textbf{92.14} & 77.14 \\
      & 100\%
      & 91.05 & 70.28
      & 91.92 & 67.79
      & 91.66 & \textbf{77.79} \\

    \bottomrule[1.5pt]
  \end{tabular}
  \end{adjustbox}
\end{table}

%% file: tables/appendix_syn_data_help_ssl.tex
\begin{table}[!t]
  \centering
  \caption{\textbf{Downstream utility of generated signals.}
    With ECGs synthesized by a text-to-ECG generator, we evaluate their utility in self-supervised learning. We found that adding synthetic data improves zero-shot transferability in ECG-text representation learning.}
  \label{tab:synthetic_data_utility}
  \begin{adjustbox}{width=0.72\linewidth}
  \footnotesize
  \setlength{\tabcolsep}{3pt}
  \begin{tabular}{l c *{6}{r}}
    \toprule[1.5pt]
    \multirow{2.5}{*}{\textbf{Setting}} &
    \multirow{2.5}{*}{\textbf{+ Syn. Data}} &
    \multicolumn{2}{c}{\textbf{MI}} &
    \multicolumn{2}{c}{\textbf{STTC}} &
    \multicolumn{2}{c}{\textbf{CD}} \\
    \cmidrule(lr){3-4}\cmidrule(lr){5-6}\cmidrule(lr){7-8}
    & & AUC$^\uparrow$ & AUPRC$^\uparrow$
      & AUC$^\uparrow$ & AUPRC$^\uparrow$
      & AUC$^\uparrow$ & AUPRC$^\uparrow$ \\
    \midrule
    \midrule

    \multirow{2}{*}{\textbf{Zero-shot}}
      & \xmark
      & 67.90 & 46.03
      & 60.45 & 33.10
      & 68.18 & \textbf{48.97} \\
      & \checkmark
      & \textbf{74.59} & \textbf{56.56}
      & \textbf{73.97} & \textbf{50.89}
      & \textbf{69.40} & 43.24 \\

    \midrule

    \multirow{2}{*}{\textbf{Finetune}}
      & \xmark
      & 84.47 & \textbf{68.65}
      & 90.64 & 75.96
      & 86.13 & \textbf{76.41} \\
      & \checkmark
      & \textbf{84.56} & 68.05
      & \textbf{90.86} & \textbf{76.44}
      & \textbf{86.81} & 76.37 \\

    \bottomrule[1.5pt]
  \end{tabular}
  \end{adjustbox}
\end{table}

%% file: tables/appendix_signal_normalization.tex
\begin{table}[!t]
  \centering
  \caption{\textbf{Comparison on normalization strategies of input signals.} We report results on invasive blood pressure reconstruction. Range based normalization provides the best performance.}
  \label{tab:normalization_ablation_vitaldb}
  \begin{adjustbox}{width=0.72\linewidth}
  \footnotesize
  \setlength{\tabcolsep}{3pt}
  \begin{tabular}{l l *{5}{r}}
    \toprule[1.5pt]
    \multirow{2.5}{*}{\textbf{Model}} &
    \multirow{2.5}{*}{\textbf{Normalization}} &
    \multicolumn{5}{c}{\textbf{Sample Level}} \\
    \cmidrule(lr){3-7}
    & & MSE$^\downarrow$ & MAE$^\downarrow$ & SMSE$^\downarrow$ & PSNR$^\uparrow$ & SSIM$^\uparrow$ \\
    \midrule
    \midrule

    \multirow{2}{*}{\dit}
      & z-score
      & 0.195 & 0.311 & \textbf{0.043} & \textbf{15.76} & 0.394 \\
      & min-max $\left[-1,1\right]$
      & \textbf{0.176} & \textbf{0.283} & 0.045 & 13.56 & \textbf{0.445} \\

    \midrule

    \multirow{2}{*}{\sit}
      & z-score
      & 0.477 & 0.545 & 0.112 & 9.63 & 0.114 \\
      & min-max $\left[-1,1\right]$
      & \textbf{0.161} & \textbf{0.274} & \textbf{0.038} & \textbf{13.96} & \textbf{0.438} \\

    \bottomrule[1.5pt]
  \end{tabular}
  \end{adjustbox}
\end{table}

%% file: tables/appendix_seq_better.tex
\begin{table*}[!t]
  \centering
  \caption{\textbf{Comparison of long-sequence generation.}
  Models perform better on imputation and translation tasks when signal-side context is available.}
  \label{tab:long_sequence_evaluation}
  \vspace{-2mm}
  \begin{adjustbox}{width=\textwidth}
  \footnotesize
  \setlength{\tabcolsep}{3pt}
  \begin{tabular}{l l *{4}{r} *{4}{r}}
    \toprule[1.5pt]
    \multirow{2}{*}{\textbf{Model}} &
    \multirow{2}{*}{\textbf{Context Length}} &
    \multicolumn{4}{c}{\textbf{Sleep Signal Imputation}} &
    \multicolumn{4}{c}{\textbf{PPG-to-ECG}} \\
    \cmidrule(lr){3-6}\cmidrule(lr){7-10}
    & & MSE$^\downarrow$ & MAE$^\downarrow$ & RMSE$^\downarrow$ & PSNR$^\uparrow$
      & MSE$^\downarrow$ & MAE$^\downarrow$ & RMSE$^\downarrow$ & PSNR$^\uparrow$ \\
    \midrule
    \midrule
    
    \multirow{2}{*}{\dit}
      & Short
      & 0.286 & 0.389 & 0.512 & 12.13
      & 0.228 & 0.363 & 0.478 & 12.43 \\
      & Long
      & \textbf{0.174} & \textbf{0.260} & \textbf{0.358} & \textbf{15.99}
      & 0.209 & 0.359 & 0.439 & 13.52 \\

    \midrule

    \multirow{2}{*}{\sit}
      & Short
      & 0.270 & 0.381 & 0.502 & 12.25
      & 0.216 & 0.352 & 0.448 & 13.26 \\
      & Long
      & 0.196 & 0.278 & 0.381 & 15.46
      & \textbf{0.157} & \textbf{0.301} & \textbf{0.382} & \textbf{14.58} \\

    \bottomrule[1.5pt]
  \end{tabular}
  \end{adjustbox}
  \vspace{-3mm}
\end{table*}

%% file: tables/appendix_seq_worse.tex
\begin{table}[!t]
  \centering
  \caption{\textbf{Comparison of long-sequence generation.} Models struggle with long signals in semantic-to-signal generation.}
  \label{tab:capture24_long_sequence_distribution}
  \begin{adjustbox}{width=0.6\linewidth}
  \footnotesize
  \setlength{\tabcolsep}{3pt}
  \begin{tabular}{l l *{3}{r}}
    \toprule[1.5pt]
    \multirow{2.5}{*}{\textbf{Model}} &
    \multirow{2.5}{*}{\textbf{Context}} &
    \multicolumn{3}{c}{\textbf{Distribution Metric}} \\
    \cmidrule(lr){3-5}
    & & FID$^\downarrow$ & Precision$^\uparrow$ & Recall$^\uparrow$ \\
    \midrule
    \midrule

    \multirow{2}{*}{\dit}
      & Activity-to-IMU 30s  & 0.32 & 0.890 & 0.618 \\
      & Activity-to-IMU 300s & 0.93 & 0.813 & \textbf{0.657} \\

    \midrule

    \multirow{2}{*}{\sit}
      & Activity-to-IMU 30s  & \textbf{0.25} & \textbf{0.893} & \textbf{0.683} \\
      & Activity-to-IMU 300s & \textbf{0.41} & \textbf{0.856} & 0.641 \\

    \bottomrule[1.5pt]
  \end{tabular}
  \end{adjustbox}
\end{table}

%% file: tables/appendix_metric.tex
\begin{table}[!t]
  \centering
  \caption{\textbf{Comparisons of different levels of metrics.} Feature space metrics are more related to signal fidelity. Here we report the cardiac-signal forecasting results.}
  \label{tab:ppgdalia_forecasting_multi_space}
  \begin{adjustbox}{width=0.72\linewidth}
  \footnotesize
  \setlength{\tabcolsep}{3pt}
  \begin{tabular}{l *{2}{r} *{2}{r} r}
    \toprule[1.5pt]
    \multirow{2.5}{*}{\textbf{Model}} &
    \multicolumn{2}{c}{\textbf{Raw Waveform Level}} &
    \multicolumn{2}{c}{\textbf{Feature Space Level}} &
    \multicolumn{1}{c}{\textbf{Signal Fidelity}} \\
    \cmidrule(lr){2-3}\cmidrule(lr){4-5}\cmidrule(lr){6-6}
    & MSE$^\downarrow$ & MAE$^\downarrow$
    & MSE$^\downarrow$ & MAE$^\downarrow$
    & FID$^\downarrow$ \\
    \midrule
    \midrule

    \dit
      & 0.224 & 0.372 & 0.0085 & \textbf{0.0637} & \textbf{4.14} \\

    \sit
      & 0.203 & 0.345 & \textbf{0.0080} & 0.0661 & 5.00 \\

    \fractalgen
      & 0.206 & 0.344 & 0.0216 & 0.1010 & 15.96 \\

    \tarflow
      & 0.223 & 0.370 & 0.0111 & 0.0732 & 7.16 \\

    \mar
      & 0.198 & 0.340 & 0.0093 & 0.0681 & 6.05 \\

    \midrule
    \textsc{Mean}
      & \textbf{0.193} & \textbf{0.299} & 0.0242 & 0.1032 & 18.90 \\

    \bottomrule[1.5pt]
  \end{tabular}
  \end{adjustbox}
\end{table}

%% file: tables/appendix_presetting_modelspec.tex
\begin{table}[ht]
\scriptsize
\centering
\setlength{\tabcolsep}{3.2pt}
\renewcommand{\arraystretch}{1.25}
\caption{\textbf{Model capacity and training configuration in SensorGen.}
We instantiate each representative model with a comparable parameter scale whenever possible,
while preserving its native architectural design and optimization convention.}
\label{tab:model_capacity_settings}

\newcommand{\params}[1]{\mbox{\(\approx\!#1\mathrm{M}\)}}
\newcommand{\yeshier}{\(\checkmark\)}
\newcommand{\nohier}{\(\times\)}

\begin{tabularx}{\textwidth}{
@{}l
>{\centering\arraybackslash}p{1.5cm}
>{\centering\arraybackslash}p{1.5cm}
>{\centering\arraybackslash}p{1.5cm}
>{\centering\arraybackslash}p{1.5cm}
>{\centering\arraybackslash}p{1.5cm}
>{\centering\arraybackslash}X@{}}
    \toprule[1.5pt]
    \textbf{Model}
    & \textbf{Total Params}
    & \textbf{Hier.}
    & \textbf{Model width}
    & \textbf{EMA}
    & \textbf{LR}
    & \textbf{Warmup} \\
    \midrule\midrule

    DiT
    & \params{272}
    & \nohier
    & 768
    & 0.9998
    & \(10^{-4}\)
    & -- \\[0.45em]

    SiT
    & \params{272}
    & \nohier
    & 768
    & 0.9998
    & \(10^{-4}\)
    & -- \\[0.45em]

    TarFlow
    & \params{143}
    & \nohier
    & 512
    & --
    & \(10^{-4}\)
    & ~\(10\%\) of total steps \\[0.45em]

    MAR
    & \params{338}
    & \nohier
    & 768
    & 0.9999
    & \(10^{-4}\)
    & ~\(10\%\) of total epochs \\[0.45em]

    FractalGen
    & \params{441}
    & \yeshier
    & 1024
    & --
    & \(5 \times 10^{-5}\)
    & ~\(10\%\) of total epochs \\[0.45em]

    Imagen
    & \params{547}
    & \yeshier
    & 768
    & --
    & \(10^{-4}\)
    & ~\(10\%\) of total steps \\

    \bottomrule[1.5pt]
\end{tabularx}

\vspace{0.25em}

\end{table}

%% file: tables/appendix_presetting_tasks.tex
\begin{table}[ht]
    \scriptsize
    \centering
    \setlength{\tabcolsep}{2.6pt}
    \renewcommand{\arraystretch}{1.18}
    \caption{\textbf{Configurations of generation target and input conditions.}}
    \label{tab:task_pretrain_settings}

    \begin{tabularx}{\textwidth}{
    @{}p{1.55cm}
    p{1.35cm}
    p{1.8cm}
    >{\raggedright\arraybackslash}X
    >{\raggedright\arraybackslash}X
    >{\raggedright\arraybackslash}X
    >{\centering\arraybackslash}p{0.95cm}
    >{\centering\arraybackslash}p{0.75cm}@{}}
        \toprule[1.5pt]
          \textbf{Dataset}
        & \textbf{Task}
        & \textbf{Target \(x\)}
        & \textbf{Semantic condition \(c_1\)}
        & \textbf{Signal-side condition \(c_2\)}
        & \textbf{Window \(T\)}
        & \textbf{Span} \\
        \midrule\midrule

        MIMIC-IV ECG
        & Text-to-ECG
        & 12-lead ECG
        & Free-text ECG report
        & --
        & \(1{,}000\)
        & 10 s \\[0.45em]

        PhyMER
        & Emotion-to-EEG
        & 14-channel EEG
        & Emotion label
        & --
        & \(1{,}280\)
        & 5 s \\[0.45em]

        CAPTURE-24
        & Activity-to-IMU
        & 3-axis accelerometer
        & Activity label
        & --
        & \(900\)
        & 30 s \\[0.6em]

        PPG-DaLiA
        & Forecasting
        & Future HR, ECG, and BVP
        & --
        & Historical HR, ECG, and BVP
        & \(6{,}000\)
        & 60 s \\[0.45em]

        SHHS
        & Sleep imputation
        & Masked middle segment
        & --
        & Left/right context from EEG, EOG, ECG, and EMG
        & \(1{,}920\)
        & 30 s \\[0.45em]

        VitalDB
        & Medication-aware forecasting
        & Future physiological signals
        & Medication trajectories
        & Historical physiological signals
        & \(3{,}000\)
        & 30 s \\[0.6em]

        PPG-DaLiA
        & PPG-to-ECG
        & ECG
        & --
        & PPG signal
        & \(6{,}000\)
        & 60 s \\[0.45em]

        PhyMER
        & Peripheral-to-EEG
        & 14-channel EEG
        & --
        & Temperature, EDA, and BVP
        & \(1{,}280\)
        & 5 s \\[0.45em]

        VitalDB
        & BP reconstruction
        & Arterial blood pressure
        & Non-invasive BP summary when used
        & PPG signal
        & \(1{,}500\)
        & 30 s \\[0.45em]

        Metabonet
        & CGM translation
        & Continuous glucose monitoring signal
        & Demographic covariates
        & Insulin and carbohydrate histories
        & \(2{,}016\)
        & 7 d \\[0.45em]

        Metabonet
        & CGM super-resolution
        & High-resolution CGM trajectory
        & Demographic covariates
        & Low-resolution CGM, insulin, and carbohydrate histories
        & \(2{,}016\)
        & 7 d \\[0.6em]

        MIMIC-IV ECG
        & ECG denoising
        & Clean ECG
        & denoise
        & Noisy ECG
        & 1000
        & 10 s \\[0.45em]

        MIMIC-IV ECG
        & ECG editing
        & edited ECG
        & editing guidance
        & source ECG
        & 1000
        & 10 s \\[0.45em]
        
        CAPTURE-24
        & IMU editing
        & Edited IMU signal
        & Target activity or editing guidance
        & Source IMU signal
        & 900
        & 30 s \\

        \bottomrule[1.5pt]
    \end{tabularx}

    \vspace{0.25em}
    \begin{flushleft}
    \footnotesize
    \end{flushleft}
\end{table}